\documentclass[fleqn,10pt]{wlscirep}

\usepackage[utf8]{inputenc}
\usepackage[T1]{fontenc}
\usepackage{amsmath} 
\usepackage{svg} 
\usepackage{float}
\usepackage[section]{placeins}
\title{Learning from pandemics: using extraordinary events can improve disease now-casting models }

\author[1]{Sara Mesquita+}
\author[3]{Cláudio Haupt Vieira+}
\author[1]{Lília Perfeito}
\author[1,2,3,4*]{Joana Gonçalves-Sá}

\affil[1]{Social Physics and Complexity Lab - SPAC, LIP,  Avenida Prof. Gama Pinto, 1600-078 Lisboa, Portugal}
\affil[2]{Physics Department, Avenida Rovisco Pais, Instituto Superior Técnico, 1049-001, Lisboa, Portugal}
\affil[3]{Nova School of Business and Economics, Rua da Holanda, 2775-405 Carcavelos, Portugal}
\affil[4]{Instituto Gulbenkian de Ciência, Rua da Quinta Grande, 2780-156 Oeiras, Portugal}
\affil[*]{joana.gsa@tecnico.ulisboa.pt}

\affil[+]{these authors contributed equally to this work}


\begin{abstract}

Online searches have been used to study different health-related behaviours, including monitoring disease outbreaks. An obvious caveat is that several reasons can motivate individuals to seek online information and models that are blind to people's motivations are of limited use and can even mislead. This is particularly true during extraordinary public health crisis, such as the ongoing pandemic, when fear, curiosity and many other reasons can lead individuals to search for health-related information, masking the disease-driven searches.
However, health crisis can also offer an opportunity to disentangle between different drivers and learn about human behavior. 
Here, we focus on the two pandemics of the 21st century (2009-H1N1 flu and Covid-19) and propose a methodology to discriminate between search patterns linked to general information seeking (media driven) and search patterns possibly more associated with actual infection (disease driven). 
We show  that by learning from such pandemic periods, with high anxiety and media hype, it is possible to select online searches and improve model performance both in pandemic and seasonal settings. Moreover, and despite the common claim that more data is always better, our results indicate that lower volume of the right data can be better than including large volumes of apparently similar data, especially in the long run. 
Our work provides a general framework that can be applied beyond specific events and diseases, and argues that algorithms can be improved simply by using less (better) data. This has important consequences, for example, to solve the accuracy-explainability trade-off in machine-learning.
\end{abstract}
\begin{document}

\flushbottom
\maketitle
%\thispagestyle{empty}

%\noindent Please note: Abbreviations should be introduced at the first mention in the main text – no abbreviations lists. Suggested structure of main text (not enforced) is provided below.

\section*{Introduction}

Infectious diseases pose great health risks to human populations worldwide. To mitigate these risks, public health institutions have set up surveillance systems that attempt to rapidly and accurately detect disease outbreaks. These systems typically include sentinel doctors and testing labs, and enable a timely response which can limit and even stop outbreaks. However, even when in place, detection and mitigation mechanisms can fail, leading to epidemics and, more rarely, pandemics, as we are currently experiencing.
In fact, disease surveillance mechanisms that only rely on highly trained personnel, are typically expensive, limited, and slow. It has been extensively argued that these should be complemented with "Digital Era" tools, such as online information, mobility patterns, or digital contact-tracing\cite{hay2013big,ferretti2020quantifying,salathe2012digital}. Online behaviours, such as searches on Google, have proven to be very relevant tools, as health information seeking is a prevalent habit of online users \cite{fox2006online}. This methodology has been applied to follow other epidemics, such as Dengue \cite{chan2011using, althouse2011prediction, husnayain2019correlation}, Avian Influenza \cite{mollema2015disease}, and Zika surveillance \cite{teng2017dynamic}. In the case of Influenza, a very common infectious disease, the potential of online-based surveillance methods gained large support with the launch of Google Flu Trends (GFT), in 2008 \cite{GFT_site}. GFT attempted to predict the timing and magnitude of influenza activity by aggregating flu-related search trends and, contrary to traditional surveillance methods, provided reports in near real-time\cite{ginsberg2009detecting}, without the need for data on clinical visits and lab reports. More recently, many others have found strong evidence that the collective search activity of flu-infected individuals seeking health information online provides a representative signal of flu activity \cite{hickmann2015forecasting,lamb2013separating,santillana2015combining,sharpe2016evaluating,won2017early}. 
However, flu infection is not the sole (and perhaps not even the strongest) motivation for individuals to seek flu-related information online \cite{lazer2014parable}. This is particularly true during extraordinary times, such as pandemics, when it is reasonable to expect individuals to have various degrees of interest, ranging from curiosity to fear, to actual disease \cite{towers2015mass}. In fact, the GFT model missed the first wave of the 2009 flu pandemic and overestimated the magnitude of the severe 2013 seasonal flu outbreak, in the USA\cite{lazer2014parable,olson2013reassessing}. This led many authors to suggest that high media activity can lead to abnormal Google search trends, possibly leading to estimation errors  \cite{copeland2013google,lazer2014parable,funk2015nine,shih2008media,collinson2014modelling,collinson2015effects}. This "media effect" was also observed by others studying Zika \cite{tizzoni2020impact,dillard2020fear,yang2018understanding}, and contributed to the disenchantment with the potential of such tools, particularly during such "extraordinary times".

However, if we could decouple searches mostly driven by media, anxiety, or curiosity, from the ones related with actual disease, we could not only improve disease monitoring, we could also deepen our understanding of online human behavior. In the case of Google search trends, identifying what terms are more correlated with media exposure and reducing their influence in the model is crucial to correct past errors.

In this paper, we propose that the characteristics that make pandemics unique and hard to now-cast, such as media hype, can also be used as opportunities for two main reasons: 1) as pandemics tend to exacerbate behaviors, the noise (media) is of the same order of magnitude as the signal (cases), making it more visible, allowing us to discriminate between the two; and 2) because information seeking becomes less common as the pandemic progresses \cite{tausczik2012public,towers2015mass} and these different dynamics can be used when selecting the search terms. In fact, instead of ignoring pandemic periods, studying what happens during the worst possible moment can help us understand which are the search-terms more associated with the disease and the ones that were prompted by media exposure. This solution might avoid over-fitting and enable the predictive model to be more robust over time, especially during seasonal events. Therefore, we focus on the only two XXI century WHO declared pandemics and aim at learning from pandemics to now-cast seasonal epidemics (or secondary waves of the same pandemic), and improving current models by incorporating insights from information-seeking behavior.

The first pandemic of the XXI century was caused by an Influenza A(H1N1)09pdm strain (pH1N1), which emerged in Mexico in February 2009 \cite{mena2016origins}. By June 2009, pH1N1 had spread globally with around 30 000 confirmed cases in 74 countries. In most countries pH1N1 displayed a bi-phasic activity: a spring-summer wave and a fall-winter wave \cite{brammer2011surveillance,devaux2010initial}. The fall-winter wave was overall more severe than the spring-summer wave as it coincided with the common flu season (in the Northern Hemisphere), that typically provides optimal conditions for flu transmission \cite{shaman2009absolute}. The pandemic was officially declared to be over in August 2010 and a total of 18 449 laboratory-confirmed pH1N1 attributable deaths were counted (WHO, 2009). This number was later revised and pH1N1 associated mortality is now believed to have been 15 times higher than the original official number \cite{dawood2012estimated}.
The second pandemic of this century, was caused by the SARS-CoV-2 virus, first identified in the last day of 2019 in the Chinese province of Wuhan. To date, Covid-19 has infected more than 78 million people and killed more than 1,7 million people worldwide. % (CITAR https://covid19.who.int/table). These two pandemics present some similarities, but they also present some significant differences that need to be taken into consideration.

Both Covid-19 and influenza viruses cause respiratory diseases with manifestations ranging from asymptomatic or mild to severe disease and death. They share a range of symptoms and trigger similar public health measures due to common transmission mechanisms. Both pandemics have a led to a great surge in media reports and public attention across many platforms, from traditional to online social media. However, there are several differences between the two pandemics: there is still a lot of uncertainty and lack of knowledge surrounding the  SARS-CoV-2 virus, including its lethality (although it is certain to be higher than the flu for older age-groups), whether it displays seasonal behaviour, its transmission frequency and patterns, whether infection confers lasting immunity, or what are its long-term health effects, respiratory or not\cite{carlson2020misconceptions,greenhalgh2020management,arunachalam2020systems,del2020long,kanzawa2020will}. Moreover, the Covid-19 pandemic led to unique public health measures and what might be considered the largest lockdown in history, with authorities implementing several preventive measures from social distancing to isolating entire countries. These restrictions have been instrumental in reducing the impact of the pandemic, but most decision-makers acknowledged the need to loosen the confinement measures. In the interest of economic and social needs, several countries re-opened schools and businesses, and many experienced surges in cases and deaths\cite{WHOcovid19}, often referred to as second and even third waves.
At this point, and as vaccines start to be distributed mostly in developed countries, all tools that can help us in identifying outbreaks are of utmost importance and different countries are deploying  different measures such as conditional movement and contact tracing apps. 

For all these reasons, improving fast, online surveillance is even more crucial now than it was in 2009, and there are already several studies on using online data to explain and forecast Covid-19 dynamics\cite{kogan2020early,dewhurst2020divergent,liu2020machine,ayyoubzadeh2020predicting,lu2020internet,effenberger2020association}. However, and despite its potential, separating what is media hype from reporting of actual disease cases (be it on Google, Facebook, or any other platform), and understanding their impact on collective attention, has been considered a huge challenge. One of the main reasons is that the patterns are intertwined with the actual spread of a disease within a population.
Therefore, we learn from the 2009 flu pandemic and propose a system to reduce the signal to noise ratio on online-searches and now-cast the current Covid-19 pandemic. The 2009 influenza offers a great case study as it was extensively researched: precise signals of pandemic flu infections were obtained through large-scale laboratory confirmations \cite{panning2009detection}, several studies analyzed the media’s behaviour during the pandemic \cite{duncan2009media, klemm2016swine, reintjes}, including the collection of news pieces and news counts, and as the pandemic emerged at a period of widespread Internet usage \cite{seybert2010internet}, several online datasets are available (including the collective behaviour of millions of users through their search trends on Google). 
Building on these datasets and by adding insights from human behaviour, we apply our framework to the current Covid-19 pandemic and provide a robust and possibly generalizable system.

\section*{Results}

\subsection*{Dynamics of media reports and online searches do not match disease cases}

\begin{figure}[ht]
    \centering
    \includegraphics[width=\textwidth]{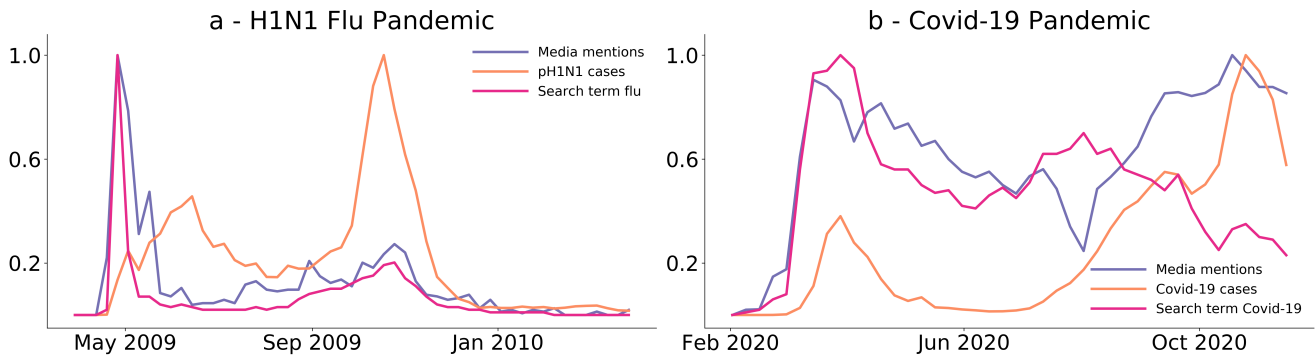}
    \caption{\textbf{Flu and Covid-19 cases during the 2009 and 2020 pandemics in US (a) and Spain (b), respectively.} \textbf{a -} Normalized weekly cases of flu (orange), media mentions (purple), and Google-trends searches for the term "flu" (pink) in the United States of America from March 2009 to March 2010. \textbf{b -} Normalized weekly cases of Covid-19 (orange), media mentions (purple), and Google-trends searches for the term "Covid-19" (pink)  between February and November 2020. All datasets are normalized to their highest value in the period. We can see a quick increase in media activity in both situations that precedes the number of cases of infection. In both panels, searches for the terms 'flu' or 'Covid-19', display a pattern more similar to the media activity trend (Pearson correlation between the search term and media of 0.85 for the flu pandemic and 0.44 for Covid-19 pandemic, compared to 0.27 and -0.03 between the search term and cases of infection, respectively).}
    \label{fig:fig1}
\end{figure}

Improving signal (disease) to noise ratio is fundamental in disease surveillance. As extraordinary events, such as pandemics, tend to become the dominant story nearly everywhere, fear and curiosity can increase and so do searches for information. First we asked whether there is a correspondence between the number of cases (for both the 2009 flu and Covid-19), media reports, and searches on Google for disease-related terms (flu and Covid-19, respectively). We focused on the US in the case of the 2009 flu and Spain during the Covid-19 pandemic. These are countries that had a large number of cases, good data availability and, in the case of Spain, already a strong second Covid-19 wave, as detailed in the methods. Figure \ref{fig:fig1} shows the number of confirmed infections, news mentions and GT searches in the United States for the 2009 pandemic (a) and in Spain for the current one (b). Since news now travel faster than pathogenic agents, the news peak for the 09 flu pandemic (figure \ref{fig:fig1}a) had its peak on the last week of April, while the first peak in cases happened later, at the end of June. More relevant is that by the time H1N1 infections had its highest peak in the US (in October/November, during regular flu season), the frequency of online searches for "flu" and news mentions had significantly reduced. In the case of the Covid-19 pandemic (figure \ref{fig:fig1}b), the early news mentions began in late 2019 when the disease was identified in China, but the first cases in Spain were only identified in February 2020 (for a similar analysis on the US case see the supplementary materials). As observed before, there was a disconnect between the intensity of the disease and both its visibility in media and the volume of Google searches\cite{lazer2014parable,olson2013reassessing}, raising the important question of whether we can discriminate between different drivers of online searches.\\

\subsection*{Online searches have different patterns}

\begin{figure}[ht]
    \centering
    \includegraphics[width=\textwidth]{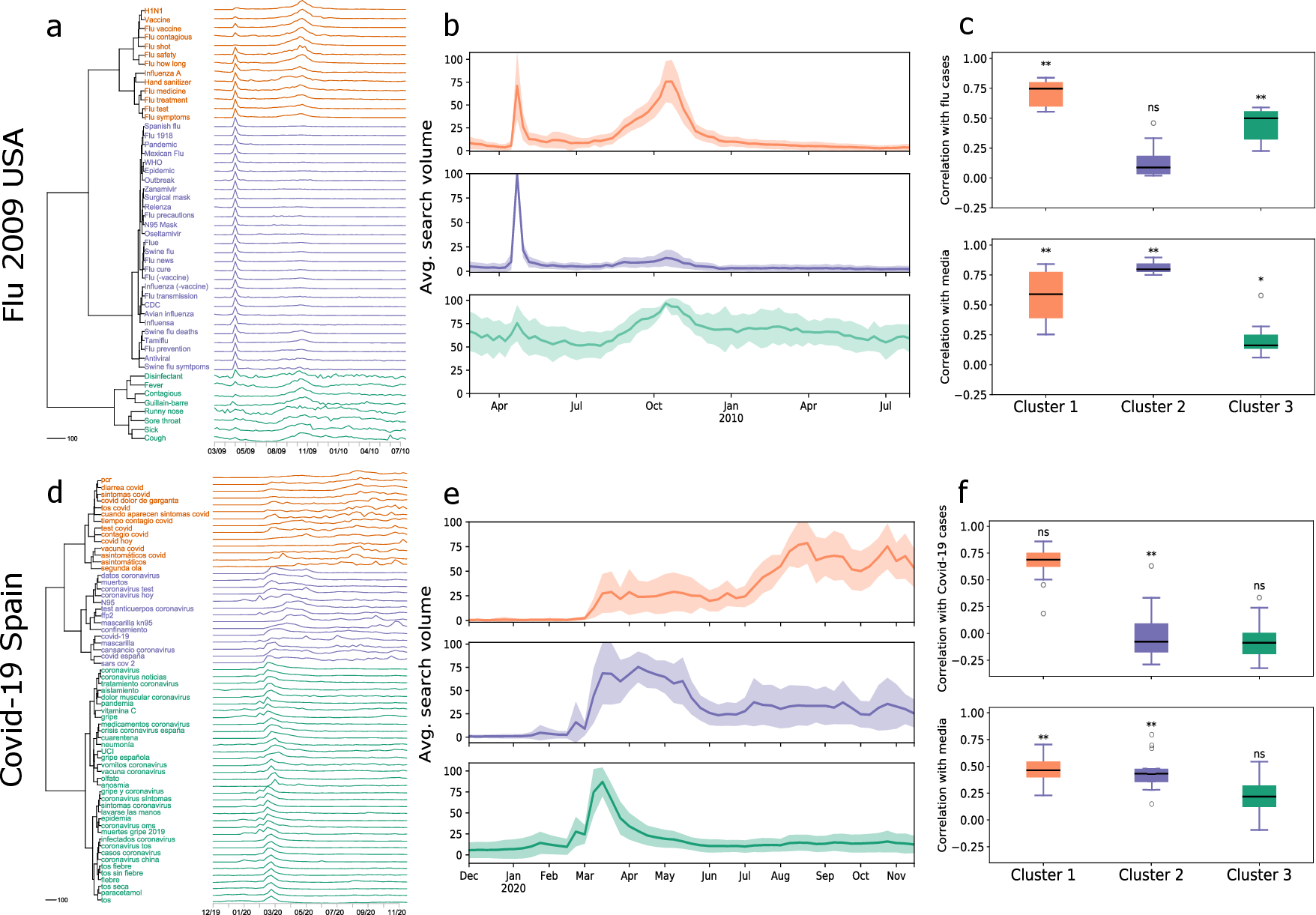}
    %\includesvg[width=\textwidth]{Figure2_071220_.svg}
    \caption{\textbf{Different patterns of searches during pandemics} Top panels refer to the 2009 flu pandemic in the USA, bottom panels refer to the COVID-19 pandemic in Spain. \textbf{a -} Dendrogram summarizing the hierarchical clustering of Google Trends search terms for the flu pandemic in US. Three clusters are very salient. \textbf{b -} Centroid and standard deviation over time for each cluster. The cluster colors correspond to the clusters in a. \textbf{c -} Pearson correlation between the cluster centroid and either the flu cases (top) or the media mentions (bottom).\mbox{*} denotes 0.01 < \textit{p-value} < 0.05, \mbox{**} denotes \textit{p-value} < 0.001, and \textit{ns} a non significant \textit{p-value}. \textbf{d -} Dendrogram summarizing the hierarchical clustering of Google Trends search terms for Covid-19 in Spain. \textbf{e -} Shows the centroid and standard deviation over time for each cluster. The cluster colors correspond to the clusters in d. \textbf{f -} Pearson correlation between the cluster centroid and either the Covid-19 cases (top) or the media mentions (bottom).}
    \label{fig:fig2}
\end{figure}

Given that the searches for "flu" and "Covid-19" do not closely follow the variation in the number of confirmed cases, we asked if we could identify particular search terms, with higher correlation with the disease progression. We started by selecting a large number of search terms, related to each disease (see supplementary materials for the full list), all of which could be \textit{a priori} considered useful for now-casting. Using hierarchical clustering, we identified three distinct clusters in both the 2009 flu and COVID-19 (\ref{fig:fig2}a and \ref{fig:fig2}d). Figures \ref{fig:fig2}b and \ref{fig:fig2}e show the centroids of each cluster, revealing the existence of different dynamics. In the case of the flu in the US, one cluster has a strong peak in the second half of 2009, another has the strongest (almost unique) peak in the first half, and a third cluster has much less clearly defined peaks (figure \ref{fig:fig2}b). The first cluster (orange) shows a strong correlation with the number of pH1N1 confirmed cases (\textit{r} = 0.78, $p = 4 \times 10^{-16}$) and a lower correlation with media (\textit{r} = 0.60, $p = 2 \times 10^{-8}$), while the second cluster (purple) has the opposite trend (figure \ref{fig:fig2}c, \textit{r} = 0.16, $p = 0.2$ with pH1N1 cases and \textit{r} = 0.83, $p = 3 \times 10^{-20}$ with media). The third has an intermediate correlation with the flu cases and poor with the media reports. As an additional test, we asked whether there was evidence that cases or media preceded any of the clusters. We performed a Granger causality test and show that that media precedes cluster 2 but not cluster 1 (supplementary materials). Neither cases nor media showed significant results for clusters 1 or 3. The grouping of the search terms is not intuitive from their meaning. Interestingly, there is no clear pattern on the search-terms that could have indicated that some would be more correlated with cases or media attention. For example, symptoms such as 'fever' or 'cough', appear on cluster 3, together with 'Guillaume-Barré syndrome' and disinfectant', while cluster 1 contains 'vaccine' and 'treatment' along with the strain of the virus and 'hand sanitizer'.  

In the case of Covid-19, the clusters are not so well defined, as shown by the smaller relative length of the internal branches of the clustering dendrogram (figure \ref{fig:fig2}d). This is likely due to a) the smaller time-frame considered (roughly half of that of H1N1 - figure \ref{fig:fig1}) b), the lower search volume, explained by the much smaller population of Spain when compared to the US, and c) the real-time nature of the analysis. Still, we could identify three clear clusters and a very similar pattern (figure \ref{fig:fig2}e): the first cluster (again orange) shows two broad peaks, the second larger than the first. The second cluster (purple) shows a clear first plateau, between March and May 2020, and the third cluster (green) a much sharper peak, encompassing little over one month. When we repeated the correlation analysis, we again identified a cluster (C1, orange) that strongly correlates with the number of cases (\textit{r} = 0.71, $p = 8 \times 10^{-6}$) but less with the media (\textit{r} = 0.52, $p = 0.003$), and a cluster (C2, purple) with the opposite pattern (a correlation with cases of \textit{r} = 0.13, $p = 0.45$ and with media of \textit{r} = 0.71, $p =  2 \times 10^{-6}$ ) (figure \ref{fig:fig2}f). Cluster 3 (green) correlates poorly with both the number of confirmed cases and media attention. 
Thus, and despite the strong entanglement and time-coincidence between the cases and the media, particularly in the case of the current pandemic, these results show that 1) not all pandemic-related search trends show the same patterns, and 2) some of the patterns may be driven by media attention whereas others by the number of cases.
%In fact, and in both cases, Granger causality shows that media reports precede the peak in cluster 2 and the number of cases precede the first peak in cluster 1 (see supplementary materials).

\subsection*{Pandemic search-terms can be used to improve seasonal forecasting}

\begin{figure}[H]
\centering
\includegraphics[width=\textwidth]{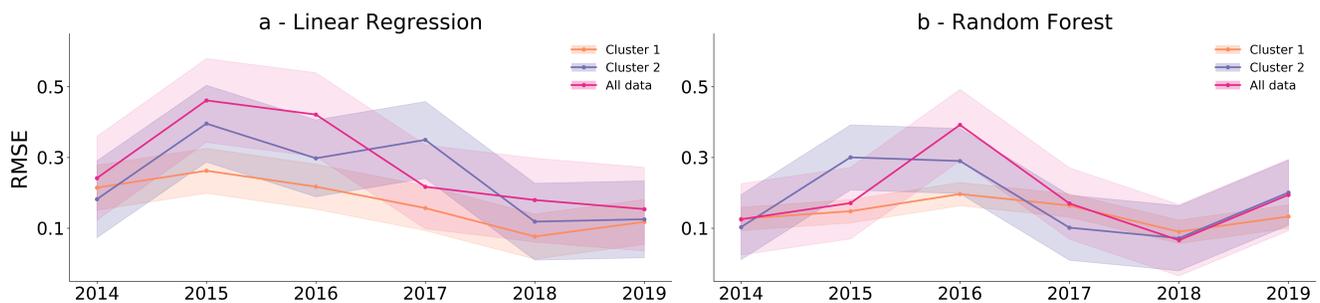}
\caption{\textbf{Performance comparison of model predictions for the flu pandemic.} \textbf{a} shows the mean squared error for the linear regression  and \textbf{b} for the Random Forest  model. Both use Google search terms from figure \ref{fig:fig2}a as independent variables to predict the seasonal flu cases between 2014 and 2019. Each dot represents the squared difference between the prediction and the empirical data, averaged over one season. Cluster 1 (orange) shows better results in almost all seasons and has a smaller standard deviation (shaded area) when compared to cluster 2 (purple) or all data (pink). In both cases, three years were used as training and the models were tested on the following year, in a sliding window process.}
\label{fig:fig3}
%\captionof{table}{Overall results for both models.}
%\begin{tabular}{@{}rrrrcrrrcrr@{}}
%& \multicolumn{6}{c}{Flu}\\
%\toprule
%& \multicolumn{3}{c}{Linear Regression} & \phantom{ab}& \multicolumn{3}{c}{Random Forest} &
%\phantom{ab}\\
%\cmidrule{2-4} \cmidrule{6-8} 
%& R\textsuperscript{2} & RMSE  &&& R\textsuperscript{2} & RMSE \\ \midrule
%cluster 1 & 0.83 &  0.17  &&& 0.86 & 0.14  \\
%cluster 2 & 0.76& 0.25 &&& 0.82& 0.18\\
%All data & 0.72& 0.28 &&& 0.81& 0.19\\
%\end{tabular}
%\label{tab:examplefl}
\end{figure}

That very similar search-terms display such different time patterns is interesting in itself but only useful if they have predictive power. Therefore, we asked whether the search terms identified as correlating with the number of confirmed cases (during a pandemic) could be used to forecast seasonal epidemics. The rationale is that if we can reduce the noise caused by the media coverage and identify the terms that are more resilient to outside factors, we can make seasonal forecasting more robust. Therefore, our goal was not to devise the best possible model, but rather to test whether particular search terms perform better than others. To do this, we took advantage of extensively available seasonal flu data and chose two simple models: a linear regression and the non-linear random forest (details in the Methods). We then tested the predictive power of the models when we used all search terms from figure \ref{fig:fig2}A (that we call "All data") or just the terms from the identified clusters in figure \ref{fig:fig2}b. For both models and all dataset variations, we used three years of data to predict the fourth and assessed the performance of the model only on the prediction season (see Methods for details). %For example, we used data from September 2004 to September 2009 to predict Sep 2010 to Sep 2011, 2005/2006 to 2010/2011 to predict 2011/2012, and so on.  
Figure \ref{fig:fig3} and table \ref{tab:examplefl} show the performance of the two models (Figure \ref{fig:fig3}a and \ref{fig:fig3}b) measured by the root-mean-square error (RME) and the coefficient of determination, R\textsuperscript{2}. In general, both models perform similarly, with a mean R\textsuperscript{2} above 0.7. In both cases, using all data (pink line) is not better than just using the terms more correlated with the number of cases during the pandemic (cluster 1, orange line), and on average cluster 1 performs better than all terms in both the linear regression (R\textsuperscript{2} = 0.81 for cluster 1 \textit{vs} R\textsuperscript{2} = 0.71 for all data) and random forest(R\textsuperscript{2} = 0.86 for cluster 1 \textit{vs} R\textsuperscript{2} = 0.81 for all data). 
%When we restrict the model to the 5 search terms most correlated with cases in cluster 1, performance improves in both models. Importantly, if we were to use the top 5 terms overall, performance would be worse than all terms (supplemental information). This means that the clustering improves the selection of terms over a simple correlation.
%, and in some seasons such as 2015/2016, just using the search-terms from cluster 1 is clearly better (Figure \ref{fig:fig3} C and D and supplemental figure 1). 
It can also be observed that cluster 1 terms (orange) tend to have a more consistent performance %, particularly in the case of the linear regression
(shown by the smaller standard deviation: $\hat{\sigma}$ = 0.08 for cluster 1 in the case of linear regression and $\hat{\sigma}$ = 0.06 for random forest  \textit{vs} $\hat{\sigma}$ = 0.163 in the case of linear regression and $\hat{\sigma}$ = 0.104 for random forest  when considering all data). 

%In fact, the random forest model, even with all data, does not significantly improve the predictions.%, but it has more variability and even wrongly predicts a peak around August??? 2012, that the simpler linear model ignores. Conversely, using the terms from cluster 3 offers the worst predictions and the largest variability, in both models. 
It is important to note that some of the features from clusters 2 and 3 might be better local predictors, and that can explain the performance of the models when using all search terms, but overall, using only the pre-identified terms of cluster 1 is better.

This indicates that 1) insights from pandemics can be used in seasonal forecasting models, and 2) refining the search-term selection, by selecting the ones less sensitive to media hype, might reduce over-fitting and improve model robustness.\\

\subsection*{Improving a model for Covid-19}

We then asked whether these results could be used in the current pandemic. This is a more challenging setting for several reasons: first, the data is arriving in close to real-time and with varying quality (the number of tests, the criteria for testing, and the reporting formats have been changing with time, even for the same country); second, there is no indication that Covid-19 might become a seasonal disease and the periodicity of new outbreaks, if any, remains unknown; third, reporting is now happening in many different online platforms, at an even faster pace than in 2009, and more importantly fourth, we do not have a large number of past seasons to train our models on. 
Still, we employed a similar approach to test whether the rationale of the flu pandemic could be applied to Covid-19. The US pandemic situation has been particular, with different states having widely different infection rates and risk levels\cite{chande2020real}. Also, at the time of this study, there were no states with clear strong second waves or evidence of seasonality. Therefore, we focused on Spain, one of the first countries to have a clear and strong second wave and trained the models on the first (February-June) wave to try to now-cast the second (June-November) wave. Still, data for the US can be found in the supplementary materials with results very consistent with what we observed in the case of Spain. 
Figure \ref{fig:fig4} shows that, again, using only the features from cluster 1 (orange) offers a much better prediction than using the search-terms from clusters 2 (purple) or 3 (supplementary materials), despite the fact that cluster 1 has a much smaller number of terms. The result is particularly striking in the case of the random forest (figure \ref{fig:fig4}b, compare pink and orange). These results further support the idea that by selecting online data, using a semi-manual approach, it is possible to improve disease now-casting.

\begin{figure}[!ht]
\centering
\includegraphics[width=\textwidth]{ModelsResult3Seasons_RSE_251120_SpainCovid.eps}
\caption{\textbf{Performance comparison of model predictions for Covid-19.} \textbf{a} shows the mean squared error for the linear regression  and \textbf{b} for the random forest  model. Both use Google search terms from figure \ref{fig:fig2}d as independent variables to predict the second wave (June to November) of Covid-19 in Spain. Each dot shows the squared difference between the prediction and the empirical data in each week. Cluster 1 (orange) presents better results in almost all seasons and has a smaller standard deviation (shaded area) when compared to cluster 2 (purple) or all data (pink). In both cases, the first wave was used to train the model.}
\label{fig:fig4}
\captionof{table}{Model results for both pandemics.}
%\begin{tabular}{@{}rrrrcrrrcrr@{}}
%& \multicolumn{6}{c}{Flu}\\
%\toprule
%& \multicolumn{3}{c}{Linear Regression} & \phantom{ab}& \multicolumn{3}{c}{Random Forest} &
%\phantom{ab}\\
%\cmidrule{2-4} \cmidrule{6-8} 
%& R\textsuperscript{2} & RMSE  &&& R\textsuperscript{2} & RMSE \\ \midrule
%cluster 1 & 0. &  0.  &&& 0.44 & 0.56  \\
%cluster 2 & 0.& 0. &&& -0.49 & 1\\
%All data & 0.& 0. &&& 0.04& 0.95\\
%\end{tabular}
%\begin{tabular}{@{} *6l @{}}    \toprule
%\begin{tabular}{@{}rrrrcrrrcrrrcrrrcrrrcrrrcrrr@{}}
\begin{tabular}{@{}crrrcrrrcrrrccc@{}}
%& \multicolumn{6}{c}{Flu}\\
%\toprule

& \multicolumn{4}{c}{Flu} 
& \multicolumn{4}{c}{Covid-19} \\
\cmidrule(lr){2-5} 
\cmidrule(lr){6-9} 

& \multicolumn{2}{c}{L. Regression} 
& \multicolumn{2}{c}{Random Forest} 
& \multicolumn{2}{c}{L. Regression} 
& \multicolumn{2}{c}{Random Forest} \\
%& \phantom{ab}& {Random Forest} 
%\phantom{ab} & {Linear Regression} & \phantom{ab} & {Random Forest} &
%\phantom{ab}\\
\cmidrule(lr){2-3} 
\cmidrule(lr){4-5} 
\cmidrule(lr){6-7} 
\cmidrule(lr){8-9}

& R\textsuperscript{2} & RMSE  & R\textsuperscript{2} & RMSE  & R\textsuperscript{2} & RMSE  & R\textsuperscript{2} & RMSE\\
\midrule
Cluster 1 & 0.83 &  0.17  & 0.86 & 0.14 & 0.96 & 0.04 & 0.84 & 0.16 \\
Cluster 2 & 0.76& 0.25 & 0.82& 0.18& 0.70 & 0.30 & 0.20 & 0.80\\
All data & 0.72& 0.28 & 0.81& 0.19& 0.46 & 0.54 & 0.35 & 0.65\\
\end{tabular}
\label{tab:examplefl}
\end{figure}

%\subsection*{Subsection}

%Example text under a subsection. Bulleted lists may be used where appropriate, e.g.

%\begin{itemize}
%\item First item
%\item Second item
%\end{itemize}

%Topical subheadings are allowed.

\section*{Discussion}

In the past, the inclusion of online data in surveillance systems has both improved the disease prediction ability over traditional syndromic surveillance systems, while also showing some very obvious caveats. Online-data based surveillance systems have many limitations and challenges, including noisy data, demographic bias, privacy issues, and, often, very limited prediction power. Previous approaches have assumed that if a search-term is a good predictor of cases in one year, it will be a good predictor in the following years \cite{ginsberg2009detecting,cook2011assessing}, when in fact, search terms may be associated with both cases and media hype in a particular year, but soon loose association with one or the other (especially when media interest fades). Moreover, and taking into consideration that these approaches often use a single explanatory-variable, meaning the model ignores the variability in individual search query tendencies over time, it can happen that terms highly correlated with disease cases in a certain moment can be highly correlated with media reports as well, but over time some might lose their association with one or another. However, and despite the described limitations, there are several successful examples of using online behaviour as a proxy for "real-world" behaviour in disease settings and it is increasingly clear that such data can offer insights not limited to disease forecasting \cite{choi2012predicting,moat2014using,stephens2014cost,vosen2011forecasting,won2017early}.

Pandemics have been particularly ignored in digital now-casting because they represent (hopefully) rare events when people's behaviour changes, making forecasting even more challenging. A large part of these behavioural changes is driven by the excess media attention: people become curious and possibly afraid, and start looking for more (different?) information. This is in contrast with seasonal outbreaks where there isn't so much relative media attention, there is more common knowledge, and people's online searches might be primarily driven by actual disease. 
In general, the notions that online search-data is too noisy and that the models used have limited prediction power have led people to try to increase the type and quantity of data, or to build more complex models. However, we argue that this tension, between using the large potential of online data and the so-called "data hubris", can be balanced in the opposite direction, by including behavioural knowledge and human curation, to reduce the amount of data required, while keeping the models simple and explainable. 

In this study, we applied this approach to two pandemics and showed that, contrary to general arguments of "more data trumps smarter algorithms"\cite{flaxman2020estimating} we can use such extraordinary events to improve seasonal forecasting, and argue that lowering the volume of data can reduce over-fitting while maintaining the quality of the predictions. This was done by actively discriminating between search queries that are very sensitive to media from queries possibly more driven by symptoms. Our approach combines elements of human curation and blind statistical inference. On the one hand our initial term list is based on knowledge of the disease. On the other, the clustering algorithm is blind to the actual meaning of the terms. This leads to unusual term-pairings such as the fact that "oseltamivir" (cluster 2), a drug used to treat flu is separate from "flu treatment" (cluster 1). We can explain this separation by considering that the media is more likely to mention the name of the drug, but that sick people might not remember it. However \textit{a priori} we might not think this distinction was important. Finally, the choice of the best cluster is again based on human curation by looking at the correlations with media and cases, which we postulate are the main drivers behind search queries. In many general now-casting problems, a similar semi-automated approach is probably more fruitful than a fully automated, data-hungry methodology.
This approach can also be particularly useful in countries where data is sparse or suffers from significant bias or delays. Even within Europe, data collection and reporting have been inconsistent, limiting global epidemiological analysis\cite{flaxman2020estimating}. Methods as the one we describe here cannot replace the need for strong, centralized, data collection systems (through the European, American or other CDCs) but might help to fill existing gaps, while surveillance networks are built or reinforced.

In addition to improving now-casting models, finding different search patterns in Google Trends can offer insights into the behaviours of internet users. Specifically, by clustering search trends on a topic we can ask whether there are different motivations behind them. If there are hypotheses about what those motivations are, they can also be tested by correlating with centroids as we do here. For example, the search terms from the media-related clusters (clusters 2) could be further analyzed to discriminate which terms are more often found in newspapers versus television, offering insight into the preferred news media. This methodology opens new doorways into connecting online and offline behaviour.

Overall, we add to the ongoing work on using digital tools and online data to improve disease monitoring and propose a new tool to now-cast infectious diseases, combining statistical tools and human curation, that can prove useful in the monitoring of the current and future pandemics and epidemics. 
%.

%1) lower volume of the right data is sometimes better than including all data
%2) pandemic search trends are clustered into groups, ones more correlated with the disease and other with media activity 
%3) If we use online searches that better correlate with disease cases to seasonal predictions, model performance can be greatly improved
%4) This approach was applied to two different pandemics, showing that is not event specific  

%Google queries are but one source of information and a multi-sourced model would be ideal
%like temperature + twitter data?

%"but was weaker at country level, as it was prone to distortions induced by unbalanced media coverage and the digital divide." from "Assessing Ebola-related web search behaviour: insights and implications from an analytical study of Google Trends-based query volumes"

%Ver artigos: 

%"A review of influenza detection and prediction through social networking sites"

%"Social media based surveillance systems for healthcare using machine learning: A systematic review": 

%The inclusion of online data in surveillance systems has improved the disease prediction ability over traditional syndromic surveillance systems. However, social media based surveillance systems have many limitations and challenges, including noise, demographic bias, privacy issues, etc.

\section*{Methods} %1500 word limit.

\subsection*{Data and Sources}

\subsubsection*{Selected countries and time period.} 
Data for the 2009 pandemic was collected for the USA, from March 2009 to August 2019,  as it offered reliable data on a large number of people. This was not possible for Covid-19 as this pandemic is reaching different states at different times and second or third waves are mostly caused by surges in new states than as a nation-wide, simultaneous epidemic. %Moreover, and as we write, there were no clear second peaks in any state.
%Moreover, and as we write, there were not many states with two clear waves\cite{NYT_byState}. 
Still, supplemental text shows that three clusters are observed, one more correlated with cases than the rest.
\
\noindent Data for the Covid-19 pandemic was collected for Spain, from January 2\textsuperscript{nd} to November 15\textsuperscript{th} 2020, as it was the country with highest number of reliable second-wave cases, offering at least one training and one testing period.
%United States flu data was used primarily due to the availability of data for a long period of time 
%EXPLAIN WHY US AND SPAIN. Also the COVID-19 state/by state disease progression

\subsubsection*{Google search trends}
Data from Google search trends (GT) \cite{GT_site} was extracted from the United States and Spain both for flu and Covid-19 pandemics, through the GT API. It provides a normalized number of queries for a given keyword, time and country \cite{ginsberg2009detecting,carneiro2009google}.
Search terms were selected to cover various aspects of pandemic and seasonal flu, and Covid-19, such as symptoms, antivirals, personal care, institutions and pandemic circumstantial terms.This was done with the help of "related queries" option that Google Trends provides, returning what people also search for when they search for a specific term. Terms that contained many "zeros" interspersed with high values were indicative of low search volume and were removed. In the end we had 49 flu-related weekly search trends in the United States and 63 Covid-related terms in Spain. Time periods were December 2019 to September 2020 in the case of Spain and September 2009 to September 2019, in the case of the USA, to cover pre-pandemic, pandemic and post-pandemic periods. In the case of the US flu pandemic, search-terms were extracted for each season separately, with a season being defined as going from September 1\textsuperscript{st} to October 1\textsuperscript{st} the following year. GT time series were extracted in September 2020 in the case of Spain, and July 2020 in the case of the US. Data was binned in a weekly resolution, to match that of reported cases and remove daily variation.
\noindent Both word lists are reported in the supplemental text.

%Google's search trends algorithm is based samples of words and thus is known to change over time (refs). So, we repeated all our analysis with the two datasets but only show the one extracted in 2020. The clusters include slightly different words, but neither the choice of cluster nor the now-casting model conclusions differ from those presented here (supplemental file Data-2018.pdf). 
 % The data was then aggregated. The extraction happened only once in 2020. 

%The complete Google Trends dataset is in ... (supplemental file)

\subsubsection*{News media}

The pandemic flu, United States media dataset contains the weekly count of both TV news broadcast and print media, that mentioned "flu" or "influenza". It includes NBC, CBS, CNN, FOX and MSNBC networks, obtained from the Vanderbilt Television News Archive \cite{BP2JXU2020}, and The New York Times, from the NYT API (https://developer.nytimes.com/). 
%The German media dataset has weekly influenza-related news counts during April 2009 until April 2010, from the ARD Tagesschau TV newscast and the newspapers Frankfurter Allgemeine Zeitung, Bild, and Spiegel \cite{reintjes}.
The Covid-19 media dataset, for both the USA and Spain was obtained through Media Cloud \cite{mediacloud}, an online open-source platform containing extensive global news corpus starting in 2011. The query "Covid-19 OR Coronavirus" was used to track media coverage of the pandemic over time. It aggregated articles that had 1 keyword, the other or both. For the case of the US, we searched the collection "United States - National" (\#34412234) and "United States - State \& Local" (\#38379429), which includes 271 national and 10,457 local media sources, respectively. For Spain we used collection "Spain - National" (\#34412356) which includes 469 media sources, and Spain - State \& Local(\#38002034), including 390 media sources.%A collection of X media websites for US and Y for Germany was considered.

\subsubsection*{Infectious Disease Data}
Data of confirmed infections from both pH1N1 and SARS-CoV-2 are publicly available. For US pH1N1 cases were extracted from the CDC’s National Respiratory and Enteric Virus Surveillance System %(Flahault et al., 1998)%
\cite{flunetdata}. In the case of Covid-19 in the US, data from national and state-level cases were extracted ECDC's Our World in Data \cite{OWID} and from New York Times \cite{NYT_data}, respectively, in August 2020. In the case of Covid-19 in Spain, data was obtained from the WHO \cite{WHOcovid19}. 

\subsection*{Analysis}

%\subsubsection*{Correlation Analysis}

%We used Spearman’s correlations to test the association between online data, media activity and pH1N1 activity. However, time series data is usually dependent on time and Spearman’s correlation is more appropriate for independent variables, so it can possibly provide misleading statistical evidence of a linear relationship \emph{i.e. spurious correlation}. To rule out the possibility of spurious correlations we employed in parallel a more sensitive test, Granger Causality. A statistically significant Granger-causality result in line with statistically significant correlations provide an indication of non-spuriousness, adding robustness to the results.

%To gauge causality, we used the Granger causality test \cite{granger1969investigating}, that assumes that if
%an event A precedes an event B, then it is possible that A is causing B. (to develop further if we decide to include this analysis)

\subsubsection*{Hierarchical clustering}

Google search terms were independently extracted from Google Trends\cite{stephens2014hands}. While all search queries include a 100, not all include a zero (if there were no weeks with less than 1\% of the maximum weekly volume), so all series were re-scaled between 0 and 100. These were clustered using hierarchical clustering, computing the pairwise Euclidean distance between words and using Ward's linkage method (an agglomerative algorithm) to construct the dendrograms shown in \ref{fig:fig2}. clustering was performed in Python, using scipy.cluster.hierarchy.dendrogram \cite{Scipy_clustering}. The number of clusters was determined through visual inspection of the dendrogram. This task was performed using data from the pandemic period, which for H1N1 pandemic was between March 2009 and August 2010, and for Covid-19 from December 2019 to September 2020.

\subsubsection*{Modeling and Evaluation}
The datasets for seasonal flu were collected similarly to those of the pandemic. They are aggregated by week and seasons were defined by visual inspection, varying from season to season, over the 9 years of data. Each dataset (cases and search time series) in each season was standardized so its mean value was 0 and its standard deviation was 1. The model was trained with 3 seasons and tested with the 4\textsuperscript{th}. In the case of Covid-19 in Spain, the data was split around the week with the fewest number of cases (June). The first wave was then used to train and the second to test.

\bigskip
\textbf{Linear Regression}\\
In each case, a model of the form 

\begin{equation}
I_i= \beta_0 + \beta_1 \times W_1 + \beta_1 \times W_1 + ... +\beta_n \times W_n + \epsilon_i
\end{equation}

was trained, where $I_i$ is the number of infections in week $i$, $\beta_0$ is the intercept, $\beta_1$ to $\beta_n$ are the coefficients of each search term and $\epsilon_i$ is the error. The coefficients were estimated as to minimize the sum of the square of the errors across all weeks.  the regression was implemented in Python using sklearn.linear\_model.LinearRegression \cite{LR} with default parameters.

\bigskip
\textbf{Random Forest}\\
For each dataset, a random forest model was trained using sklearn.ensemble.RandomForestRegressor \cite{RF} implemented in Python. The hyperparameters - number of estimators, max features and max depth - were selected through cross validation using \textit{GridSearchCV} from [10,20,50,100,200,500,1000], [0.6,0.8,"auto","sqrt"] and   [2,4,5,6] respectively. 

%"Comparison of machine learning classifiers for influenza detection from emergency department free-text reports"

%Topical subheadings are allowed. Authors must ensure that their Methods section includes adequate experimental and characterization data necessary for others in the field to reproduce their work.

%\noindent LaTeX formats citations and references automatically using the bibliography records in your .bib file, which you can edit via the project menu. Use the cite command for an inline citation, e.g.  \cite{Hao:gidmaps:2014}.

%For data citations of datasets uploaded to e.g. \emph{figshare}, please use the \verb|howpublished| option in the bib entry to specify the platform and the link, as in the \verb|Hao:gidmaps:2014| example in the sample bibliography file.

\section*{Acknowledgments}
The authors would like to thank members of the SPAC lab for comments and critical reading of the manuscript. This work was partially funded by FCT grant DSAIPA/AI/0087/2018 to JGS and by PhD fellowships SFRH/BD/139322/2018 and 2020.10157.BD to CHV and SM, respectively.

\section*{Author contributions statement}

All authors participated in project conception, data analysis, and paper writing.  

\section*{Additional information}

%To include, in this order: \textbf{Accession codes} (where applicable); \textbf{Competing interests} (mandatory statement). 

%The corresponding author is responsible for submitting a \href{http://www.nature.com/srep/policies/index.html#competing}{competing interests statement} on behalf of all authors of the paper. This statement must be included in the submitted article file.

\bibliography{sample.bib}

\begin{thebibliography}{10}
\urlstyle{rm}
\expandafter\ifx\csname url\endcsname\relax
  \def\url#1{\texttt{#1}}\fi
\expandafter\ifx\csname urlprefix\endcsname\relax\def\urlprefix{URL }\fi
\expandafter\ifx\csname doiprefix\endcsname\relax\def\doiprefix{DOI: }\fi
\providecommand{\bibinfo}[2]{#2}
\providecommand{\eprint}[2][]{\url{#2}}

\bibitem{hay2013big}
\bibinfo{author}{Hay, S.~I.}, \bibinfo{author}{George, D.~B.},
  \bibinfo{author}{Moyes, C.~L.} \& \bibinfo{author}{Brownstein, J.~S.}
\newblock \bibinfo{journal}{\bibinfo{title}{Big data opportunities for global
  infectious disease surveillance}}.
\newblock {\emph{\JournalTitle{PLoS med}}} \textbf{\bibinfo{volume}{10}},
  \bibinfo{pages}{e1001413} (\bibinfo{year}{2013}).

\bibitem{ferretti2020quantifying}
\bibinfo{author}{Ferretti, L.} \emph{et~al.}
\newblock \bibinfo{journal}{\bibinfo{title}{Quantifying sars-cov-2 transmission
  suggests epidemic control with digital contact tracing}}.
\newblock {\emph{\JournalTitle{Science}}} \textbf{\bibinfo{volume}{368}}
  (\bibinfo{year}{2020}).

\bibitem{salathe2012digital}
\bibinfo{author}{Salathe, M.} \emph{et~al.}
\newblock \bibinfo{journal}{\bibinfo{title}{Digital epidemiology}}.
\newblock {\emph{\JournalTitle{PLoS Comput Biol}}}
  \textbf{\bibinfo{volume}{8}}, \bibinfo{pages}{e1002616}
  (\bibinfo{year}{2012}).

\bibitem{fox2006online}
\bibinfo{author}{Fox, S.}
\newblock \emph{\bibinfo{title}{Online health search 2006}}
  (\bibinfo{publisher}{Pew Internet \& American Life Project},
  \bibinfo{year}{2006}).

\bibitem{chan2011using}
\bibinfo{author}{Chan, E.~H.}, \bibinfo{author}{Sahai, V.},
  \bibinfo{author}{Conrad, C.} \& \bibinfo{author}{Brownstein, J.~S.}
\newblock \bibinfo{journal}{\bibinfo{title}{Using web search query data to
  monitor dengue epidemics: a new model for neglected tropical disease
  surveillance}}.
\newblock {\emph{\JournalTitle{PLoS neglected tropical diseases}}}
  \textbf{\bibinfo{volume}{5}}, \bibinfo{pages}{e1206} (\bibinfo{year}{2011}).

\bibitem{althouse2011prediction}
\bibinfo{author}{Althouse, B.~M.}, \bibinfo{author}{Ng, Y.~Y.} \&
  \bibinfo{author}{Cummings, D.~A.}
\newblock \bibinfo{journal}{\bibinfo{title}{Prediction of dengue incidence
  using search query surveillance}}.
\newblock {\emph{\JournalTitle{PLoS Negl Trop Dis}}}
  \textbf{\bibinfo{volume}{5}}, \bibinfo{pages}{e1258} (\bibinfo{year}{2011}).

\bibitem{husnayain2019correlation}
\bibinfo{author}{Husnayain, A.}, \bibinfo{author}{Fuad, A.} \&
  \bibinfo{author}{Lazuardi, L.}
\newblock \bibinfo{journal}{\bibinfo{title}{Correlation between google trends
  on dengue fever and national surveillance report in indonesia}}.
\newblock {\emph{\JournalTitle{Global Health Action}}}
  \textbf{\bibinfo{volume}{12}}, \bibinfo{pages}{1552652}
  (\bibinfo{year}{2019}).

\bibitem{mollema2015disease}
\bibinfo{author}{Mollema, L.} \emph{et~al.}
\newblock \bibinfo{journal}{\bibinfo{title}{c}}.
\newblock {\emph{\JournalTitle{Journal of medical Internet research}}}
  \textbf{\bibinfo{volume}{17}}, \bibinfo{pages}{e128} (\bibinfo{year}{2015}).

\bibitem{teng2017dynamic}
\bibinfo{author}{Teng, Y.} \emph{et~al.}
\newblock \bibinfo{journal}{\bibinfo{title}{Dynamic forecasting of zika
  epidemics using google trends}}.
\newblock {\emph{\JournalTitle{PloS one}}} \textbf{\bibinfo{volume}{12}},
  \bibinfo{pages}{e0165085} (\bibinfo{year}{2017}).

\bibitem{GFT_site}
\bibinfo{title}{Google flu trends}.
\newblock
  \bibinfo{howpublished}{\url{https://web.archive.org/web/20121022154915/http://www.google.org/flutrends/about/how.html}}.
\newblock \bibinfo{note}{Accessed: 2020-12-22}.

\bibitem{ginsberg2009detecting}
\bibinfo{author}{Ginsberg, J.} \emph{et~al.}
\newblock \bibinfo{journal}{\bibinfo{title}{Detecting influenza epidemics using
  search engine query data}}.
\newblock {\emph{\JournalTitle{Nature}}} \textbf{\bibinfo{volume}{457}},
  \bibinfo{pages}{1012--1014} (\bibinfo{year}{2009}).

\bibitem{hickmann2015forecasting}
\bibinfo{author}{Hickmann, K.~S.} \emph{et~al.}
\newblock \bibinfo{journal}{\bibinfo{title}{Forecasting the 2013--2014
  influenza season using wikipedia}}.
\newblock {\emph{\JournalTitle{PLoS Comput Biol}}}
  \textbf{\bibinfo{volume}{11}}, \bibinfo{pages}{e1004239}
  (\bibinfo{year}{2015}).

\bibitem{lamb2013separating}
\bibinfo{author}{Lamb, A.}, \bibinfo{author}{Paul, M.} \&
  \bibinfo{author}{Dredze, M.}
\newblock \bibinfo{title}{Separating fact from fear: Tracking flu infections on
  twitter}.
\newblock In \emph{\bibinfo{booktitle}{Proceedings of the 2013 Conference of
  the North American Chapter of the Association for Computational Linguistics:
  Human Language Technologies}}, \bibinfo{pages}{789--795}
  (\bibinfo{year}{2013}).

\bibitem{santillana2015combining}
\bibinfo{author}{Santillana, M.} \emph{et~al.}
\newblock \bibinfo{journal}{\bibinfo{title}{Combining search, social media, and
  traditional data sources to improve influenza surveillance}}.
\newblock {\emph{\JournalTitle{PLoS Comput Biol}}}
  \textbf{\bibinfo{volume}{11}}, \bibinfo{pages}{e1004513}
  (\bibinfo{year}{2015}).

\bibitem{sharpe2016evaluating}
\bibinfo{author}{Sharpe, J.~D.}, \bibinfo{author}{Hopkins, R.~S.},
  \bibinfo{author}{Cook, R.~L.} \& \bibinfo{author}{Striley, C.~W.}
\newblock \bibinfo{journal}{\bibinfo{title}{Evaluating google, twitter, and
  wikipedia as tools for influenza surveillance using bayesian change point
  analysis: a comparative analysis}}.
\newblock {\emph{\JournalTitle{JMIR public health and surveillance}}}
  \textbf{\bibinfo{volume}{2}}, \bibinfo{pages}{e161} (\bibinfo{year}{2016}).

\bibitem{won2017early}
\bibinfo{author}{Won, M.}, \bibinfo{author}{Marques-Pita, M.},
  \bibinfo{author}{Louro, C.} \& \bibinfo{author}{Gon{\c{c}}alves-S{\'a}, J.}
\newblock \bibinfo{journal}{\bibinfo{title}{Early and real-time detection of
  seasonal influenza onset}}.
\newblock {\emph{\JournalTitle{PLoS computational biology}}}
  \textbf{\bibinfo{volume}{13}}, \bibinfo{pages}{e1005330}
  (\bibinfo{year}{2017}).

\bibitem{lazer2014parable}
\bibinfo{author}{Lazer, D.}, \bibinfo{author}{Kennedy, R.},
  \bibinfo{author}{King, G.} \& \bibinfo{author}{Vespignani, A.}
\newblock \bibinfo{journal}{\bibinfo{title}{The parable of google flu: traps in
  big data analysis}}.
\newblock {\emph{\JournalTitle{Science}}} \textbf{\bibinfo{volume}{343}},
  \bibinfo{pages}{1203--1205} (\bibinfo{year}{2014}).

\bibitem{towers2015mass}
\bibinfo{author}{Towers, S.} \emph{et~al.}
\newblock \bibinfo{journal}{\bibinfo{title}{Mass media and the contagion of
  fear: the case of ebola in america}}.
\newblock {\emph{\JournalTitle{PloS one}}} \textbf{\bibinfo{volume}{10}},
  \bibinfo{pages}{e0129179} (\bibinfo{year}{2015}).

\bibitem{olson2013reassessing}
\bibinfo{author}{Olson, D.~R.}, \bibinfo{author}{Konty, K.~J.},
  \bibinfo{author}{Paladini, M.}, \bibinfo{author}{Viboud, C.} \&
  \bibinfo{author}{Simonsen, L.}
\newblock \bibinfo{journal}{\bibinfo{title}{Reassessing google flu trends data
  for detection of seasonal and pandemic influenza: a comparative
  epidemiological study at three geographic scales}}.
\newblock {\emph{\JournalTitle{PLoS Comput Biol}}}
  \textbf{\bibinfo{volume}{9}}, \bibinfo{pages}{e1003256}
  (\bibinfo{year}{2013}).

\bibitem{copeland2013google}
\bibinfo{author}{Copeland, P.} \emph{et~al.}
\newblock \bibinfo{title}{Google disease trends: an update}.
\newblock In \emph{\bibinfo{booktitle}{International Society of Neglected
  Tropical Diseases 2013}}, \bibinfo{pages}{3} (\bibinfo{year}{2013}).

\bibitem{funk2015nine}
\bibinfo{author}{Funk, S.} \emph{et~al.}
\newblock \bibinfo{journal}{\bibinfo{title}{Nine challenges in incorporating
  the dynamics of behaviour in infectious diseases models}}.
\newblock {\emph{\JournalTitle{Epidemics}}} \textbf{\bibinfo{volume}{10}},
  \bibinfo{pages}{21--25} (\bibinfo{year}{2015}).

\bibitem{shih2008media}
\bibinfo{author}{Shih, T.-J.}, \bibinfo{author}{Wijaya, R.} \&
  \bibinfo{author}{Brossard, D.}
\newblock \bibinfo{journal}{\bibinfo{title}{Media coverage of public health
  epidemics: Linking framing and issue attention cycle toward an integrated
  theory of print news coverage of epidemics}}.
\newblock {\emph{\JournalTitle{Mass Communication \& Society}}}
  \textbf{\bibinfo{volume}{11}}, \bibinfo{pages}{141--160}
  (\bibinfo{year}{2008}).

\bibitem{collinson2014modelling}
\bibinfo{author}{Collinson, S.} \& \bibinfo{author}{Heffernan, J.~M.}
\newblock \bibinfo{journal}{\bibinfo{title}{Modelling the effects of media
  during an influenza epidemic}}.
\newblock {\emph{\JournalTitle{BMC public health}}}
  \textbf{\bibinfo{volume}{14}}, \bibinfo{pages}{376} (\bibinfo{year}{2014}).

\bibitem{collinson2015effects}
\bibinfo{author}{Collinson, S.}, \bibinfo{author}{Khan, K.} \&
  \bibinfo{author}{Heffernan, J.~M.}
\newblock \bibinfo{journal}{\bibinfo{title}{The effects of media reports on
  disease spread and important public health measurements}}.
\newblock {\emph{\JournalTitle{PloS one}}} \textbf{\bibinfo{volume}{10}},
  \bibinfo{pages}{e0141423} (\bibinfo{year}{2015}).

\bibitem{tizzoni2020impact}
\bibinfo{author}{Tizzoni, M.}, \bibinfo{author}{Panisson, A.},
  \bibinfo{author}{Paolotti, D.} \& \bibinfo{author}{Cattuto, C.}
\newblock \bibinfo{journal}{\bibinfo{title}{The impact of news exposure on
  collective attention in the united states during the 2016 zika epidemic}}.
\newblock {\emph{\JournalTitle{PLoS computational biology}}}
  \textbf{\bibinfo{volume}{16}}, \bibinfo{pages}{e1007633}
  (\bibinfo{year}{2020}).

\bibitem{dillard2020fear}
\bibinfo{author}{Dillard, J.~P.}, \bibinfo{author}{Li, R.} \&
  \bibinfo{author}{Yang, C.}
\newblock \bibinfo{journal}{\bibinfo{title}{Fear of zika: Information seeking
  as cause and consequence}}.
\newblock {\emph{\JournalTitle{Health Communication}}} \bibinfo{pages}{1--11}
  (\bibinfo{year}{2020}).

\bibitem{yang2018understanding}
\bibinfo{author}{Yang, C.}, \bibinfo{author}{Dillard, J.~P.} \&
  \bibinfo{author}{Li, R.}
\newblock \bibinfo{journal}{\bibinfo{title}{Understanding fear of zika:
  Personal, interpersonal, and media influences}}.
\newblock {\emph{\JournalTitle{Risk Analysis}}} \textbf{\bibinfo{volume}{38}},
  \bibinfo{pages}{2535--2545} (\bibinfo{year}{2018}).

\bibitem{tausczik2012public}
\bibinfo{author}{Tausczik, Y.}, \bibinfo{author}{Faasse, K.},
  \bibinfo{author}{Pennebaker, J.~W.} \& \bibinfo{author}{Petrie, K.~J.}
\newblock \bibinfo{journal}{\bibinfo{title}{Public anxiety and information
  seeking following the h1n1 outbreak: blogs, newspaper articles, and wikipedia
  visits}}.
\newblock {\emph{\JournalTitle{Health communication}}}
  \textbf{\bibinfo{volume}{27}}, \bibinfo{pages}{179--185}
  (\bibinfo{year}{2012}).

\bibitem{mena2016origins}
\bibinfo{author}{Mena, I.} \emph{et~al.}
\newblock \bibinfo{journal}{\bibinfo{title}{Origins of the 2009 h1n1 influenza
  pandemic in swine in mexico}}.
\newblock {\emph{\JournalTitle{Elife}}} \textbf{\bibinfo{volume}{5}},
  \bibinfo{pages}{e16777} (\bibinfo{year}{2016}).

\bibitem{brammer2011surveillance}
\bibinfo{author}{Brammer, L.} \emph{et~al.}
\newblock \bibinfo{journal}{\bibinfo{title}{Surveillance for influenza during
  the 2009 influenza a (h1n1) pandemic--united states, april 2009--march
  2010}}.
\newblock {\emph{\JournalTitle{Clinical Infectious Diseases}}}
  \textbf{\bibinfo{volume}{52}}, \bibinfo{pages}{S27--S35}
  (\bibinfo{year}{2011}).

\bibitem{devaux2010initial}
\bibinfo{author}{Devaux, I.} \emph{et~al.}
\newblock \bibinfo{journal}{\bibinfo{title}{Initial surveillance of 2009
  influenza a (h1n1) pandemic in the european union and european economic area,
  april--september 2009}}.
\newblock {\emph{\JournalTitle{Eurosurveillance}}}
  \textbf{\bibinfo{volume}{15}}, \bibinfo{pages}{19740} (\bibinfo{year}{2010}).

\bibitem{shaman2009absolute}
\bibinfo{author}{Shaman, J.} \& \bibinfo{author}{Kohn, M.}
\newblock \bibinfo{journal}{\bibinfo{title}{Absolute humidity modulates
  influenza survival, transmission, and seasonality}}.
\newblock {\emph{\JournalTitle{Proceedings of the National Academy of
  Sciences}}} \textbf{\bibinfo{volume}{106}}, \bibinfo{pages}{3243--3248}
  (\bibinfo{year}{2009}).

\bibitem{dawood2012estimated}
\bibinfo{author}{Dawood, F.~S.} \emph{et~al.}
\newblock \bibinfo{journal}{\bibinfo{title}{Estimated global mortality
  associated with the first 12 months of 2009 pandemic influenza a h1n1 virus
  circulation: a modelling study}}.
\newblock {\emph{\JournalTitle{The Lancet infectious diseases}}}
  \textbf{\bibinfo{volume}{12}}, \bibinfo{pages}{687--695}
  (\bibinfo{year}{2012}).

\bibitem{carlson2020misconceptions}
\bibinfo{author}{Carlson, C.~J.}, \bibinfo{author}{Gomez, A.~C.},
  \bibinfo{author}{Bansal, S.} \& \bibinfo{author}{Ryan, S.~J.}
\newblock \bibinfo{journal}{\bibinfo{title}{Misconceptions about weather and
  seasonality must not misguide covid-19 response}}.
\newblock {\emph{\JournalTitle{Nature Communications}}}
  \textbf{\bibinfo{volume}{11}}, \bibinfo{pages}{1--4} (\bibinfo{year}{2020}).

\bibitem{greenhalgh2020management}
\bibinfo{author}{Greenhalgh, T.}, \bibinfo{author}{Knight, M.},
  \bibinfo{author}{Buxton, M.}, \bibinfo{author}{Husain, L.} \emph{et~al.}
\newblock \bibinfo{journal}{\bibinfo{title}{Management of post-acute covid-19
  in primary care}}.
\newblock {\emph{\JournalTitle{bmj}}} \textbf{\bibinfo{volume}{370}}
  (\bibinfo{year}{2020}).

\bibitem{arunachalam2020systems}
\bibinfo{author}{Arunachalam, P.~S.} \emph{et~al.}
\newblock \bibinfo{journal}{\bibinfo{title}{Systems biological assessment of
  immunity to mild versus severe covid-19 infection in humans}}.
\newblock {\emph{\JournalTitle{Science}}} \textbf{\bibinfo{volume}{369}},
  \bibinfo{pages}{1210--1220} (\bibinfo{year}{2020}).

\bibitem{del2020long}
\bibinfo{author}{Del~Rio, C.}, \bibinfo{author}{Collins, L.~F.} \&
  \bibinfo{author}{Malani, P.}
\newblock \bibinfo{journal}{\bibinfo{title}{Long-term health consequences of
  covid-19}}.
\newblock {\emph{\JournalTitle{Jama}}} \textbf{\bibinfo{volume}{324}},
  \bibinfo{pages}{1723--1724} (\bibinfo{year}{2020}).

\bibitem{kanzawa2020will}
\bibinfo{author}{Kanzawa, M.}, \bibinfo{author}{Spindler, H.},
  \bibinfo{author}{Anglemyer, A.} \& \bibinfo{author}{Rutherford, G.~W.}
\newblock \bibinfo{journal}{\bibinfo{title}{Will coronavirus disease 2019
  become seasonal?}}
\newblock {\emph{\JournalTitle{The Journal of infectious diseases}}}
  \textbf{\bibinfo{volume}{222}}, \bibinfo{pages}{719--721}
  (\bibinfo{year}{2020}).

\bibitem{WHOcovid19}
\bibinfo{title}{Who coronavirus disease}.
\newblock \bibinfo{howpublished}{\url{https://covid19.who.int/}}.
\newblock \bibinfo{note}{Accessed: 2020-10-01}.

\bibitem{kogan2020early}
\bibinfo{author}{Kogan, N.~E.} \emph{et~al.}
\newblock \bibinfo{journal}{\bibinfo{title}{An early warning approach to
  monitor covid-19 activity with multiple digital traces in near real-time}}.
\newblock {\emph{\JournalTitle{arXiv preprint arXiv:2007.00756}}}
  (\bibinfo{year}{2020}).

\bibitem{dewhurst2020divergent}
\bibinfo{author}{Dewhurst, D.~R.} \emph{et~al.}
\newblock \bibinfo{journal}{\bibinfo{title}{Divergent modes of online
  collective attention to the covid-19 pandemic are associated with future
  caseload variance}}.
\newblock {\emph{\JournalTitle{arXiv preprint arXiv:2004.03516}}}
  (\bibinfo{year}{2020}).

\bibitem{liu2020machine}
\bibinfo{author}{Liu, D.} \emph{et~al.}
\newblock \bibinfo{journal}{\bibinfo{title}{A machine learning methodology for
  real-time forecasting of the 2019-2020 covid-19 outbreak using internet
  searches, news alerts, and estimates from mechanistic models}}.
\newblock {\emph{\JournalTitle{arXiv preprint arXiv:2004.04019}}}
  (\bibinfo{year}{2020}).

\bibitem{ayyoubzadeh2020predicting}
\bibinfo{author}{Ayyoubzadeh, S.~M.}, \bibinfo{author}{Ayyoubzadeh, S.~M.},
  \bibinfo{author}{Zahedi, H.}, \bibinfo{author}{Ahmadi, M.} \&
  \bibinfo{author}{Kalhori, S. R.~N.}
\newblock \bibinfo{journal}{\bibinfo{title}{Predicting covid-19 incidence
  through analysis of google trends data in iran: data mining and deep learning
  pilot study}}.
\newblock {\emph{\JournalTitle{JMIR Public Health and Surveillance}}}
  \textbf{\bibinfo{volume}{6}}, \bibinfo{pages}{e18828} (\bibinfo{year}{2020}).

\bibitem{lu2020internet}
\bibinfo{author}{Lu, T.} \& \bibinfo{author}{Reis, B.~Y.}
\newblock \bibinfo{journal}{\bibinfo{title}{Internet search patterns reveal
  clinical course of disease progression for covid-19 and predict pandemic
  spread in 32 countries}}.
\newblock {\emph{\JournalTitle{medRxiv}}}  (\bibinfo{year}{2020}).

\bibitem{effenberger2020association}
\bibinfo{author}{Effenberger, M.} \emph{et~al.}
\newblock \bibinfo{journal}{\bibinfo{title}{Association of the covid-19
  pandemic with internet search volumes: a google trendstm analysis}}.
\newblock {\emph{\JournalTitle{International Journal of Infectious Diseases}}}
  (\bibinfo{year}{2020}).

\bibitem{panning2009detection}
\bibinfo{author}{Panning, M.} \emph{et~al.}
\newblock \bibinfo{journal}{\bibinfo{title}{Detection of influenza a (h1n1) v
  virus by real-time rt-pcr}}.
\newblock {\emph{\JournalTitle{Eurosurveillance}}}
  \textbf{\bibinfo{volume}{14}}, \bibinfo{pages}{19329} (\bibinfo{year}{2009}).

\bibitem{duncan2009media}
\bibinfo{author}{Duncan, B.}
\newblock \bibinfo{journal}{\bibinfo{title}{How the media reported the first
  days of the pandemic (h1n1) 2009: results of eu-wide media analysis}}.
\newblock {\emph{\JournalTitle{Eurosurveillance}}}
  \textbf{\bibinfo{volume}{14}}, \bibinfo{pages}{19286} (\bibinfo{year}{2009}).

\bibitem{klemm2016swine}
\bibinfo{author}{Klemm, C.}, \bibinfo{author}{Das, E.} \&
  \bibinfo{author}{Hartmann, T.}
\newblock \bibinfo{journal}{\bibinfo{title}{Swine flu and hype: a systematic
  review of media dramatization of the h1n1 influenza pandemic}}.
\newblock {\emph{\JournalTitle{Journal of Risk Research}}}
  \textbf{\bibinfo{volume}{19}}, \bibinfo{pages}{1--20} (\bibinfo{year}{2016}).

\bibitem{reintjes}
\bibinfo{author}{Reintjes, R.} \emph{et~al.}
\newblock \bibinfo{journal}{\bibinfo{title}{“pandemic public health
  paradox”: time series analysis of the 2009/10 influenza a/h1n1
  epidemiology, media attention, risk perception and public reactions in 5
  european countries}}.
\newblock {\emph{\JournalTitle{PloS one}}} \textbf{\bibinfo{volume}{11}},
  \bibinfo{pages}{e0151258} (\bibinfo{year}{2016}).

\bibitem{seybert2010internet}
\bibinfo{author}{Seybert, H.} \& \bibinfo{author}{L{\"o}{\"o}f, A.}
\newblock \bibinfo{journal}{\bibinfo{title}{Internet usage in 2010--households
  and individuals}}.
\newblock {\emph{\JournalTitle{Eurostat, data in Focus}}}
  \bibinfo{pages}{50--2010} (\bibinfo{year}{2010}).

\bibitem{chande2020real}
\bibinfo{author}{Chande, A.} \emph{et~al.}
\newblock \bibinfo{journal}{\bibinfo{title}{Real-time, interactive website for
  us-county-level covid-19 event risk assessment}}.
\newblock {\emph{\JournalTitle{Nature Human Behaviour}}}
  \textbf{\bibinfo{volume}{4}}, \bibinfo{pages}{1313--1319}
  (\bibinfo{year}{2020}).

\bibitem{cook2011assessing}
\bibinfo{author}{Cook, S.}, \bibinfo{author}{Conrad, C.},
  \bibinfo{author}{Fowlkes, A.~L.} \& \bibinfo{author}{Mohebbi, M.~H.}
\newblock \bibinfo{journal}{\bibinfo{title}{Assessing google flu trends
  performance in the united states during the 2009 influenza virus a (h1n1)
  pandemic}}.
\newblock {\emph{\JournalTitle{PloS one}}} \textbf{\bibinfo{volume}{6}},
  \bibinfo{pages}{e23610} (\bibinfo{year}{2011}).

\bibitem{choi2012predicting}
\bibinfo{author}{Choi, H.} \& \bibinfo{author}{Varian, H.}
\newblock \bibinfo{journal}{\bibinfo{title}{Predicting the present with google
  trends}}.
\newblock {\emph{\JournalTitle{Economic record}}}
  \textbf{\bibinfo{volume}{88}}, \bibinfo{pages}{2--9} (\bibinfo{year}{2012}).

\bibitem{moat2014using}
\bibinfo{author}{Moat, H.~S.}, \bibinfo{author}{Preis, T.},
  \bibinfo{author}{Olivola, C.~Y.}, \bibinfo{author}{Liu, C.} \&
  \bibinfo{author}{Chater, N.}
\newblock \bibinfo{journal}{\bibinfo{title}{Using big data to predict
  collective behavior in the real world 1}}.
\newblock {\emph{\JournalTitle{Behavioral and Brain Sciences}}}
  \textbf{\bibinfo{volume}{37}}, \bibinfo{pages}{92--93}
  (\bibinfo{year}{2014}).

\bibitem{stephens2014cost}
\bibinfo{author}{Stephens-Davidowitz, S.}
\newblock \bibinfo{journal}{\bibinfo{title}{The cost of racial animus on a
  black candidate: Evidence using google search data}}.
\newblock {\emph{\JournalTitle{Journal of Public Economics}}}
  \textbf{\bibinfo{volume}{118}}, \bibinfo{pages}{26--40}
  (\bibinfo{year}{2014}).

\bibitem{vosen2011forecasting}
\bibinfo{author}{Vosen, S.} \& \bibinfo{author}{Schmidt, T.}
\newblock \bibinfo{journal}{\bibinfo{title}{Forecasting private consumption:
  survey-based indicators vs. google trends}}.
\newblock {\emph{\JournalTitle{Journal of forecasting}}}
  \textbf{\bibinfo{volume}{30}}, \bibinfo{pages}{565--578}
  (\bibinfo{year}{2011}).

\bibitem{flaxman2020estimating}
\bibinfo{author}{Flaxman, S.} \emph{et~al.}
\newblock \bibinfo{journal}{\bibinfo{title}{Estimating the effects of
  non-pharmaceutical interventions on covid-19 in europe}}.
\newblock {\emph{\JournalTitle{Nature}}} \textbf{\bibinfo{volume}{584}},
  \bibinfo{pages}{257--261} (\bibinfo{year}{2020}).

\bibitem{GT_site}
\bibinfo{title}{Google trends}.
\newblock
  \bibinfo{howpublished}{\url{https://trends.google.com/trends/?geo=US}}.
\newblock \bibinfo{note}{Accessed: 2020-10-16}.

\bibitem{carneiro2009google}
\bibinfo{author}{Carneiro, H.~A.} \& \bibinfo{author}{Mylonakis, E.}
\newblock \bibinfo{journal}{\bibinfo{title}{Google trends: a web-based tool for
  real-time surveillance of disease outbreaks}}.
\newblock {\emph{\JournalTitle{Clinical infectious diseases}}}
  \textbf{\bibinfo{volume}{49}}, \bibinfo{pages}{1557--1564}
  (\bibinfo{year}{2009}).

\bibitem{BP2JXU2020}
\bibinfo{author}{Sood, G.} \& \bibinfo{author}{Laohaprapanon, S.}
\newblock \bibinfo{title}{{Vanderbilt TV News Abstracts}},
  \doiprefix\url{10.7910/DVN/BP2JXU} (\bibinfo{year}{2020}).

\bibitem{mediacloud}
\bibinfo{title}{Media cloud}.
\newblock \bibinfo{howpublished}{\url{https://mediacloud.org/}}.
\newblock \bibinfo{note}{Accessed: 2021-01-04}.

\bibitem{flunetdata}
\bibinfo{title}{Flunet}.
\newblock
  \bibinfo{howpublished}{\url{https://www.who.int/influenza/gisrs_laboratory/flunet/en/}}.
\newblock \bibinfo{note}{Accessed: 2020-06-18}.

\bibitem{OWID}
\bibinfo{title}{Our world in data}.
\newblock
  \bibinfo{howpublished}{\url{https://github.com/owid/covid-19-data/tree/master/public/data}}.
\newblock \bibinfo{note}{Accessed: 2020-08-20}.

\bibitem{NYT_data}
\bibinfo{title}{New york times covid-19 data}.
\newblock
  \bibinfo{howpublished}{\url{https://github.com/nytimes/covid-19-data}}.
\newblock \bibinfo{note}{Accessed: 2020-08-20}.

\bibitem{stephens2014hands}
\bibinfo{author}{Stephens-Davidowitz, S.} \& \bibinfo{author}{Varian, H.}
\newblock \bibinfo{journal}{\bibinfo{title}{A hands-on guide to google data}}.
\newblock {\emph{\JournalTitle{further details on the construction can be found
  on the Google Trends page}}}  (\bibinfo{year}{2014}).

\bibitem{Scipy_clustering}
\bibinfo{title}{Scipy clustering}.
\newblock \bibinfo{howpublished}{\url{
  https://docs.scipy.org/doc/scipy/reference/generated/scipy.cluster.hierarchy.dendrogram.html}}.

\bibitem{LR}
\bibinfo{title}{Linearregression}.
\newblock \bibinfo{howpublished}{\url{
  https://scikit-learn.org/stable/modules/generated/sklearn.linear_model.LinearRegression.htmll}}.

\bibitem{RF}
\bibinfo{title}{Randomforestregressor}.
\newblock
  \bibinfo{howpublished}{\url{https://scikit-learn.org/stable/modules/generated/sklearn.ensemble.RandomForestRegressor.html}}.

\end{thebibliography}

\end{document}

\maketitle

\section{Introduction}

This template is designed to assist with creating a supplemental document to accompany an article in an OSA journal. This template contains example content to help you create your document, and you may use this template as a visual guide. The sections below show examples of different components and styles.

\section{Numbering Items in the Supplementary Document}

The supplementary materials document may contain additional figures, tables, equations, etc. Such items should be numbered, with an uppercase “S” to identify them as supplementary. For example, number the first figure in the supplementary document “Fig. S1”; the first table “Table S1”; etc.

This template has been designed to automatically format these components with this styling, but we include the naming convention here for reference.

\subsection*{Naming Convention for Countable Items}

\begin{condenseditemize}
\item[] Algorithm S1
\item[] Equation (S1)
\item[] Figure S1
\item[] Media S1
\item[] Table S1
\end{condenseditemize}

\section{Figures and Tables}
Figures and Tables should be labeled and referenced in the standard way using the \verb|\label{}| and \verb|\ref{}| commands.

\subsection{Sample Figure}

Figure \ref{fig:false-color} shows an example figure.

\begin{figure}[htbp]
\centering
\fbox{\includegraphics[width=.6\linewidth]{sample}}
\caption{False-color image, where each pixel is assigned to one of seven reference spectra.}
\label{fig:false-color}
\end{figure}

\subsection{Sample Table}

Table \ref{tab:shape-functions} shows an example table. 

\begin{table}[htbp]
\centering
\caption{\bf Shape Functions for Quadratic Line Elements}
\begin{tabular}{ccc}
\hline
local node & $\{N\}_m$ & $\{\Phi_i\}_m$ $(i=x,y,z)$ \\
\hline
$m = 1$ & $L_1(2L_1-1)$ & $\Phi_{i1}$ \\
$m = 2$ & $L_2(2L_2-1)$ & $\Phi_{i2}$ \\
$m = 3$ & $L_3=4L_1L_2$ & $\Phi_{i3}$ \\
\hline
\end{tabular}
  \label{tab:shape-functions}
\end{table}

\section{Sample Equation}

Let $X_1, X_2, \ldots, X_n$ be a sequence of independent and identically distributed random variables with $\text{E}[X_i] = \mu$ and $\text{Var}[X_i] = \sigma^2 < \infty$, and let
\begin{equation}
S_n = \frac{X_1 + X_2 + \cdots + X_n}{n}
      = \frac{1}{n}\sum_{i}^{n} X_i
\label{eq:refname1}
\end{equation}
denote their mean. Then as $n$ approaches infinity, the random variables $\sqrt{n}(S_n - \mu)$ converge in distribution to a normal $\mathcal{N}(0, \sigma^2)$.

\section{Sample Algorithm}

Algorithms can be included using the commands as shown in algorithm \ref{alg:euclid}.

\begin{algorithm}
\caption{Euclid’s algorithm}\label{alg:euclid}
\begin{algorithmic}[1]
\Procedure{Euclid}{$a,b$}\Comment{The g.c.d. of a and b}
\State $r\gets a\bmod b$
\While{$r\not=0$}\Comment{We have the answer if r is 0}
\State $a\gets b$
\State $b\gets r$
\State $r\gets a\bmod b$
\EndWhile\label{euclidendwhile}
\State \textbf{return} $b$\Comment{The gcd is b}
\EndProcedure
\end{algorithmic}
\end{algorithm}

\section*{Media}

The supplemental document may contain linked objects such as video, 2D, 3D, and machine-readable data files. Please see the \href{https://www.opticsinfobase.org/submit/style/supplementary_materials.cfm}{Author Guidelines for Supplementary Materials} for more information. Such files should be cited in the supplementary document as in the primary document but using the naming convention described above.

\section*{References} 

The supplementary materials document may contain a reference list. The reference list should follow OSA's citation style and should be checked carefully, since no copyediting will be performed by OSA staff. You may add citations manually or use BibTeX. See \cite{Zhang:14}.

Citations that are relevant to the primary manuscript and the supplementary document may be included in both places.

% Bibliography
\bibliography{sample}

%Manual citation list
%\begin{thebibliography}{1}
%\bibitem{Zhang:14}
%Y.~Zhang, S.~Qiao, L.~Sun, Q.~W. Shi, W.~Huang, %L.~Li, and Z.~Yang,
 % \enquote{Photoinduced active terahertz metamaterials with nanostructured
  %vanadium dioxide film deposited by sol-gel method,} Opt. Express \textbf{22},
  %11070--11078 (2014).
%\end{thebibliography}

\maketitle
\tableofcontents
\section{Full List of search terms for Flu US and Covid-19 Spain}

\begin{figure}[H]
    \centering
    \includegraphics[width=\textwidth]{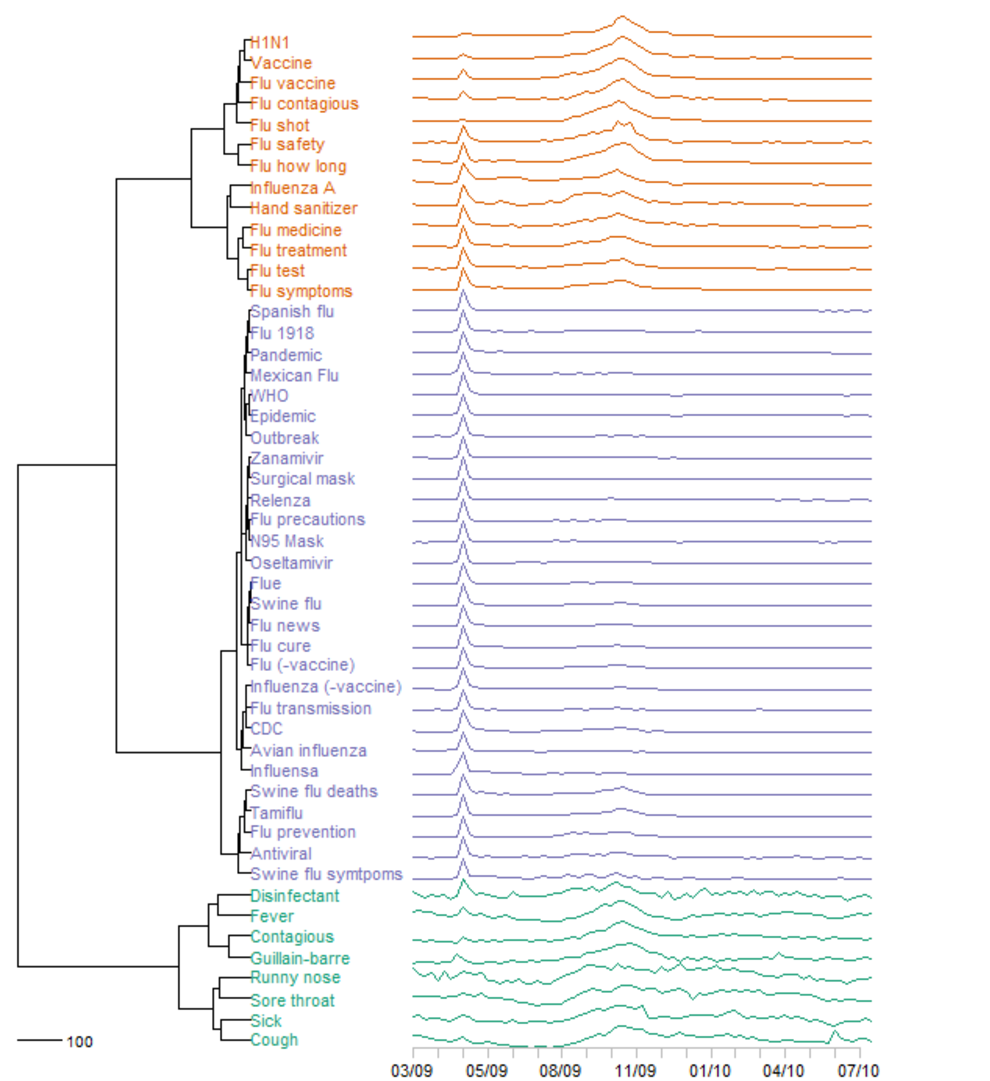}
    \caption{\textbf{Google trends search terms dendogram for US.}  The dendrogram was obtained through agglomerative hierarchical clustering using Euclidean distance and Ward’s linkage criterion. The pandemic period (March 2009-July 2010) was used and three distinct clusters were identified: Cluster 1 (orange), Cluster 2 (purple), Cluster 3 (green). }
    \label{fig:Dend_US}
\end{figure}

\textbf{Collected terms US - Flu pandemic:} 
 
\noindent H1N1, Vaccine, Flu vaccine, Flu contagious, Flu shot, Flu safety, Flu how long, Influenza A, Hand sanitizer,
Flu medicine, Flu treatment, Flu test, Flu symptoms, Spanish Flu, Flu 1918, Pandemic, Mexican Flu, WHO,
Epidemic, Outbreak, Zanamivir, Surgical mask, Relenza, Flu precautions, N95 mask, Oseltamivir, Flue, Swine flu,
Flu news, Flu cure, Flu (-vaccine), Influenza (-vaccine), Flu transmission, CDC, Avian influenza, Influensa, 
Swine flu deaths, Tamiflu, Flu Prevention, Antiviral, Swine flu symptoms, Disinfectant, Fever, Contagious, 
Guillain-barre, Runny nose, Sore throat, Sick, Cough

\begin{figure}[H]
    \centering
    \includegraphics[width=\textwidth]{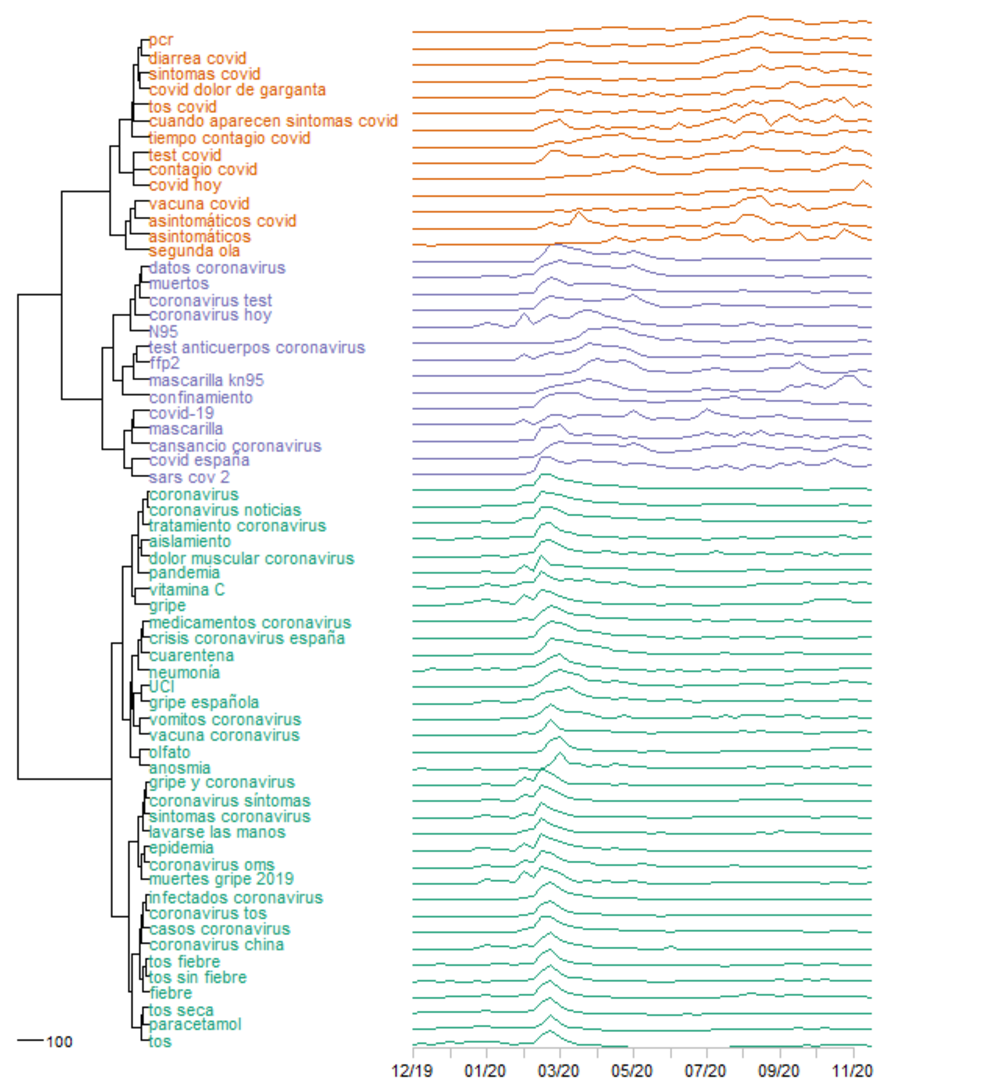}
    \caption{\textbf{Collected terms Spain(ES) dendrogram.} The dendrogram was obtained through agglomerative hierarchical clustering using Euclidean distance and Ward’s linkage criterion. The selected period to build the clusters was between December 2019 to September 2020. Three clusters were identified and search terms membership can be observed: Cluster 1 (orange), Cluster 2 (purple) and Cluster 3 (green).}
    \label{fig:ES_221120}
\end{figure}

\noindent\textbf{Collected terms ES - Covid-19 pandemic:} 
 
\noindent pcr, diarrea covid, sintomas covid, covid dolor de garganta, tos covid, cuando aparecen sintomas covid,
tiempo contagio covid, test covid, contagio covid, covid hoy, vacuna covid, asintomáticos covid, asintomáticos,
segunda ola, datos coronavirus, muertos, coronavirus test, coronavirus hoy, N95, test anticuerpos coronavirus, 
ffp2, mascarilla kn95, confinamiento, covid-19, mascarilla, cansancio coronavirus, covid españa, sars cov 2,
coronavirus, coronavirus noticias, tratamiento coronavirus, aislamiento, dolor muscular coronavirus, pandemia,
vitamina C, gripe, medicamentos coronavirus, crisis coronavirus españa, cuarentena, neumonía, UCI, gripe española,
vomitos coronavirus, vacuna coronavirus, olfato, anosmia, gripe y coronavirus, coronavirus síntomas, 
sintomas coronavirus, lavarse las manos, epidemia, coronavirus oms, muertes gripe 2019, infectados coronavirus,
coronavirus tos, casos coronavirus, coronavirus china, tos fiebre, tos sin fiebre, fiebre, tos seca, paracetamol,
tos

\section{Model results for Flu US and Covid-19 Spain}

\begin{table}[H]\centering
\begin{tabular}{@{}crrrcrrrcrrrccc@{}}
%& \multicolumn{6}{c}{Flu}\\
%\toprule

& \multicolumn{4}{c}{Flu} 
& \multicolumn{4}{c}{Covid-19} \\
\cmidrule(lr){2-5} 
\cmidrule(lr){6-9} 

& \multicolumn{2}{c}{L. Regression} 
& \multicolumn{2}{c}{Random Forest} 
& \multicolumn{2}{c}{L. Regression} 
& \multicolumn{2}{c}{Random Forest} \\
%& \phantom{ab}& {Random Forest} 
%\phantom{ab} & {Linear Regression} & \phantom{ab} & {Random Forest} &
%\phantom{ab}\\
\cmidrule(lr){2-3} 
\cmidrule(lr){4-5} 
\cmidrule(lr){6-7} 
\cmidrule(lr){8-9}

& R\textsuperscript{2} & RMSE  & R\textsuperscript{2} & RMSE  & R\textsuperscript{2} & RMSE  & R\textsuperscript{2} & RMSE\\
\midrule
Cluster 1 & 0.83 &  0.17  & 0.86 & 0.14 & 0.96 & 0.04 & 0.84 & 0.16 \\
Cluster 2 & 0.76& 0.25 & 0.82& 0.18& 0.70 & 0.30 & 0.20 & 0.80\\
Cluster 3 & 0.50 & 0.50 & 0.53 & 0.47 & 0.55 & 0.45 & 0.44 & 0.56\\
All data & 0.72& 0.28 & 0.81& 0.19& 0.46 & 0.54 & 0.35 & 0.65\\
\end{tabular}
\label{tab:examplefl}
\end{table}

We collected weekly counts of laboratory-confirmed cases of all strains of flu combined for USA.

\section {Covid-19 in the USA}
\subsection{Cases and media mention for Covid-19 in the USA}

Data from Google search trends (GT), news media and Covid-19 infection cases for the USA was extracted from January 26\textsuperscript{th} to November 15\textsuperscript{th} 2020. We used GT API to obtain the weekly search terms volume, news media was extracted using Media Cloud \cite{mediacloud} and the infection cases collected using the Covid-19 tracking API from \emph{The Atlantic}\cite{TheAtlanticAPI}.

\begin{figure}[H]
    \centering
    \includegraphics[width=\textwidth]{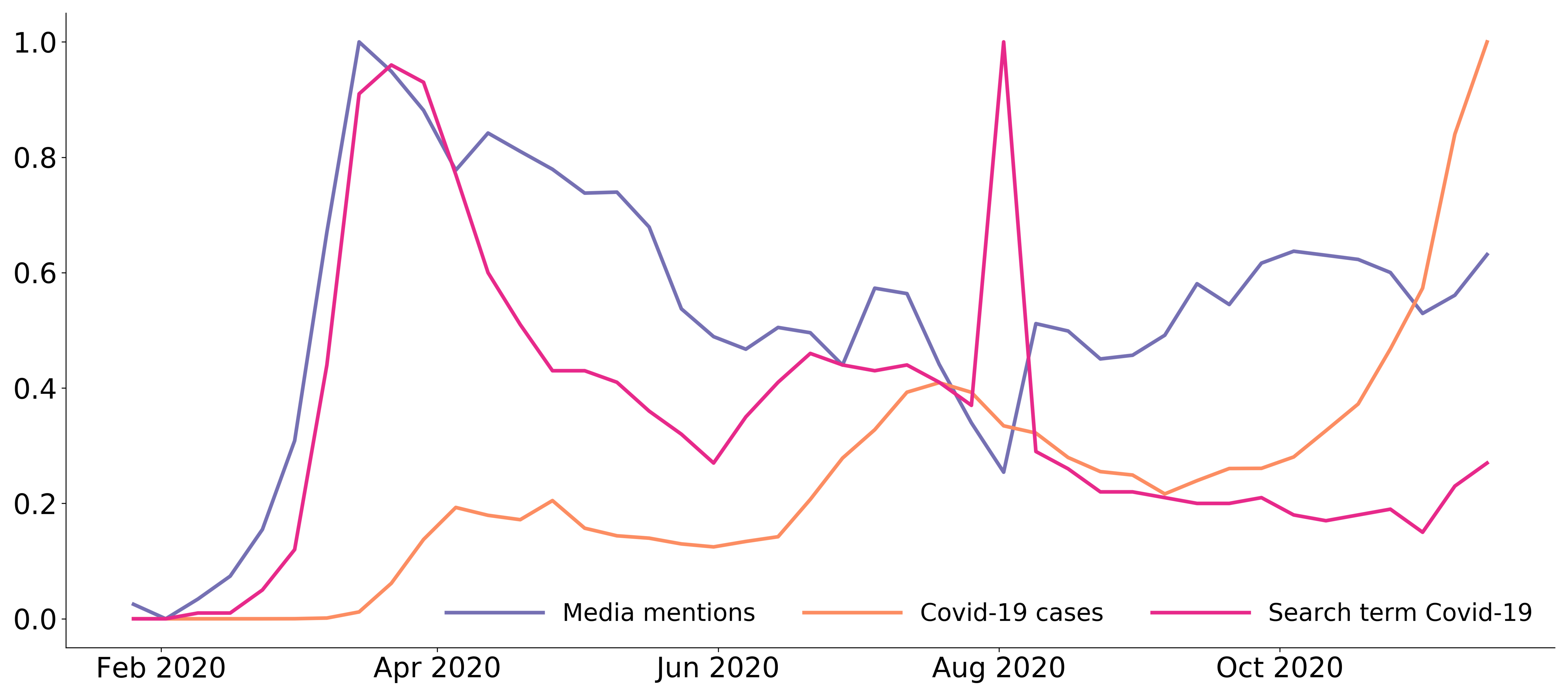}
    \caption{\textbf{Covid-19 cases during the 2020 pandemic in the US.} Weekly cases of Covid-19 (in orage), media mentions (in purple) and the search volume for the search term 'Covid-19' in the United States of America from January 2020 until mid November 2020. We can see a quick increase in media activity that precedes the number of cases of infection. Also, some search terms, like 'Covid-19', have very similar apparent trends to media activity.}
    \label{fig:Figure_S3_Suppl}
\end{figure}

\subsection{Search Terms}
\
\\ \noindent\textbf{Collected terms US - Covid-19 pandemic:}\\

\noindent covid-19 testing, covid-19 symptoms, hot spots, covid pneumonia, covid cough, chest pain covid, covid symptoms,
covid, loss taste, covid cold symptoms, vaccine, pcr, pcr test, flu shot, stay home, anosmia, 
when will quarantine end, nk95, contact tracing, covid-19 news, icu beds, icu, loss taste and smell, 
loss of smell, covid anosmia, quarantine, isolation, coronavirus cases, covid-19 map, coronavirus covid-19, 
pandemic, coronavirus deaths, spanish flu, flu 2019, flu deaths, deaths, ffp2, covid virus, covid-19 cases, 
chest pain, nhs, corona, dry cough, coronavirus, cdc, outbreak, epidemic, coronavirus symptoms, sore throat,     	
fever, cold symptoms, coronavirus symptoms vs cold, cold flu symptoms, flu symptoms, influenza, pneumonia symptoms,
cough and fever, respiratory infection, pneumonia, cough, flu, oseltamivir tamiflu, oseltamivir phosphate,
rsv, oseltamivir, tamiflu, symptoms influenza, bronchitis contagious, robitussin, bronchitis pneumonia,
bronchitis
\subsection{Clusters}
\
\\ \noindent\ We chose to identify the search terms that best characterize search behavior across the United States. This selection was used to predict new cases across the country. To better understand whether we could also predict new infections locally, we selected three states that had two well-defined waves of infections (train with one wave to test on the second wave). However, the search terms included in each state were not always the same, as terms containing zeros were discarded.

\begin{figure}[H]
    \centering
    \includegraphics[width=\textwidth]{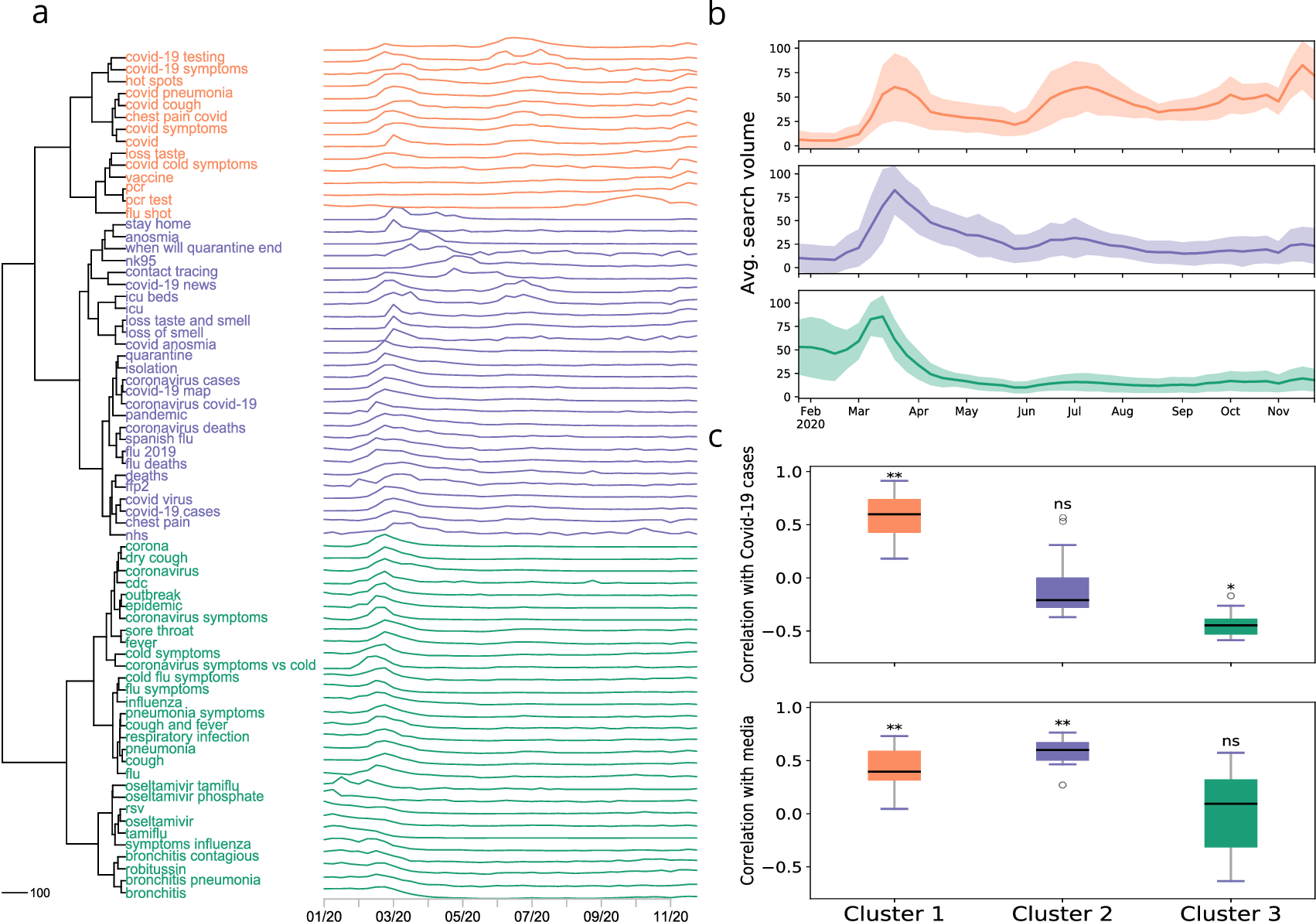}
    %\includesvg[width=\textwidth]{US_global_171220.svg}
    \caption{\textbf{Different patterns of searches during the Covid-19 pandemic from January to November 2020 in the USA.} \textbf{a -} Dendrogram summarizing the hierarchical clustering of Google Trends search terms for the Covid-19 pandemic in US. Three clusters are very salient. \textbf{b -} Centroid and standard deviation over time for each cluster. The cluster colors correspond to the clusters in a. \textbf{c -} Pearson correlation between the cluster centroid and either the Covid-19 cases (top) or the media mentions (bottom).\mbox{*} denotes 0.01 < \textit{p-value} < 0.05, \mbox{**} denotes \textit{p-value} < 0.001, and \textit{ns} a non significant \textit{p-value}.}
    \label{fig:US_global_171220}
\end{figure}

\subsection{Nowcasting}
%Models were trained with the first wave of Covid-19 cases for all the USA and for each state, independently. The second wave was used to test model predictions. From the results shown in Table \ref{tab:resultsUScovid} we can see that better predictions were obtained for All USA and Arizona. This can reflect the fact that search terms were selected considering all country (ISO 3166-2:US) and not individual states. Also, among the selected states, Arizona is the state that has two clear waves of similar magnitude. This suggests that our method is robust but would return better results if we considered each state as an independent "country".%

Models were trained with the first wave of Covid-19 cases for all the USA, and the second wave was used to test model predictions. From the results shown in Table \ref{tab:resultsUScovid} we can see that better predictions were obtained with Cluster 1, the one more correlated with cases, in both models.

% The period considered for all USA predictions was between 2020-01-21 and 2020-11-26. The first wave used to train starts from 2020-03-08 and goes until 2020-09-06. The second wave used to train started from 2020-09-13 until 2020-11-15.  %Only the random forest is shown because the linear regression models were all very poor (negative $R^2$).
%\begin{table}[H]\centering
%\begin{tabular}{@{}crrrcrrrcrrrccc@{}}
%& \multicolumn{2}{c}{1st Wave} 
%& \multicolumn{2}{c}{2nd Wave} \\
%\midrule
%All USA & 2020-03-08 to 2020-09-06  &&& 2020-09-13 to 2020-11-15 \\
%Arizona & 2020-03-08 to 2020-09-06 &&& 2020-09-13 to 2020-11-22\\
%Texas & 2020-03-08 to 2020-10-11 &&& 2020-10-18 to 2020-11-22\\
%Florida & 2020-03-08 to 2020-10-04 &&& 2020-10-11 to 2020-11-22\\
%\end{tabular}
%\caption{Dates considered for the first and second waves of Covid-19 pandemic.}
%\end{table}

\begin{table}[H]\centering
\ra{1.3}
\begin{tabular}{@{}rccccc@{}} %rrrrcrrrcrrr
\multicolumn{6}{c}{Covid-19 USA}\\
\toprule
& \multicolumn{2}{c}{Random Forest} & \phantom{ab}& \multicolumn{2}{c}{Linear Regression} \\
%&
%\phantom{abc}\\
\cmidrule{2-3} \cmidrule{5-6} 
& R\textsuperscript{2} & RMSE && R\textsuperscript{2} & RMSE \\ \midrule
All data & 0.63 &  0.37  && 0.35 & 0.65 \\
Cluster 1 & 0.85 &  0.15  && 0.87 & 0.13 \\
Cluster 2 & -0.6 & 1.06 && -25.29 & 26.29\\
Cluster 3 & -0.46 & 1.46 && -130.99 & 131.99\\

%\hline
%\multicolumn{6}{c}{Arizona}\\
%\toprule
%& \multicolumn{2}{c}{Random Forest} & \phantom{ab}& \multicolumn{2}{c}{Linear Regression} \\
%\phantom{abc}\\
%\cmidrule{2-3} \cmidrule{5-6}
%& R\textsuperscript{2} & RSE && R\textsuperscript{2} & RSE \\ \midrule
%All data & 0.80 &  0.20  && 0.49 & 0.51\\
%Cluster 1 & 0.86 &  0.14  && 0.81 & 0.19\\
%Cluster 2 & -0.02 & 1.02 && 0.58 & 0.42\\
%Cluster 3 & 0.83 & 0.17 && -33.6 & 34.6\\

%\hline
%\multicolumn{6}{c}{Texas}\\
%\toprule
%& \multicolumn{2}{c}{Random Forest} & \phantom{ab}& %\multicolumn{2}{c}{Linear Regression} \\
%\cmidrule{2-3} \cmidrule{5-6}
%& R\textsuperscript{2} & RSE && R\textsuperscript{2} & RSE \\ %\midrule
%All data & 0.72 &  0.28  && -4.53 & 5.53 \\
%Cluster 1 & 0.70 &  0.30  && -2.02 & 3.02 \\
%Cluster 2 & -0.17 & 1.17 && -23.70 & 24.70\\
%Cluster 3 & -0.81 & 1.81 && 0.35 & 0.65\\

%\hline
%\multicolumn{6}{c}{Florida}\\
%\toprule
%& \multicolumn{2}{c}{Random Forest} & \phantom{ab}& %\multicolumn{2}{c}{Linear Regression} \\
%\cmidrule{2-3} \cmidrule{5-6} 
%& R\textsuperscript{2} & RSE && R\textsuperscript{2} & RSE \\ \midrule
%All data & 0.73 &  0.27  && 0.05 & 0.95 \\
%Cluster 1 & 0.69 &  0.31  && -33.76 & 34.76 \\
%Cluster 2 & 0.31 & 0.69 && -3.51 & 4.51\\
%Cluster 3 & -0.27 & 1.27 && 0.37 & 0.63\\

\bottomrule
\end{tabular}

\caption{Predictions results obtained using linear regression and random forest models for the Covid-19 pandemic in the USA. }
\label{tab:resultsUScovid}
\end{table}

\section {Granger causality}

To test whether there was evidence of causality between media, the news and each of the clusters, a lagged regression model was fitted to the data:

\begin{equation}
C (t)= \sum_{\tau=1}^{L} b_\tau \times x_i(t + \tau) + \epsilon_{i}
\end{equation}
where C(t) are the autoregressed values of the cluster centroid at week $t$ (autoregression was performed using the function \textit{diff} in R \cite{R} with order 1), $b_\tau$ are the fitted regression coefficients, $x_i(t)$ are the values of the time series to be tested at week $t$ and $\epsilon_i$ are the residuals. $L$ is the number of lags tested, which was chosen as the one explaining more variance in $C$, using the function \textit{VARselect} \cite{vars}. Lags up to four weeks were tried. Granger causality was tested by computing $G = \log (\frac{\sigma_j}{\sigma_i})$, where $\sigma_j$ and $\sigma_i$ are the variance of the residuals in the autoregression and in the lagged regression, respectively. $G$ is expected to follow a $\chi^2$ distribution with $L$ degrees of freedom. The test was implemented using \textit{grangertest} \cite{lmtest}.

\begin{table}[H]\centering
\ra{1.3}
\begin{tabular}{@{}rrrrcrrrcrrr@{}}
& \multicolumn{6}{c}{Flu}\\
\toprule
& \multicolumn{3}{c}{Cases} & \phantom{abc}& \multicolumn{3}{c}{Media} &
\phantom{abc}\\
\cmidrule{2-4} \cmidrule{6-8} \cmidrule{10-12}
& Order & $G$ statistic & $p$ value && Order & $G$ statistic & $p$ value\\ \midrule
Cluster 1 & 2 &  1.95 & 0.15 && 4 & 1.94 & 0.11 \\
Cluster 2 & 3& 0.52& 0.67&& 4& 11.95& $\mathbf{3 \times 10^{-7}}$\\
Cluster 3 & 2& 1.91& 0.16&& 45& 1.21& 0.31\\

\hline
& \multicolumn{6}{c}{Covid-19 Spain}\\
\toprule
& \multicolumn{3}{c}{Cases} & \phantom{abc}& \multicolumn{3}{c}{Media} &
\phantom{abc}\\
\cmidrule{2-4} \cmidrule{6-8} \cmidrule{10-12}
& Order & $G$ statistic & $p$ value && Order & $G$ statistic & $p$ value\\ \midrule
Cluster 1 & 1 & 0.03 & 0.86 && 1 &0.15 & 0.70 \\
Cluster 2 & 1& 0.95& 0.34&& 1& 0.03& 0.86\\
Cluster 3& 2& 0.76& 0.48&& 1& 0.79& 0.38\\

%\hline
%& \multicolumn{6}{c}{Covid-19 Arizona}\\
%\toprule
%& \multicolumn{3}{c}{Cases} & \phantom{abc}& %\multicolumn{3}{c}{Media} &
%\phantom{abc}\\
%\cmidrule{2-4} \cmidrule{6-8} \cmidrule{10-12}
%& Order & $G$ statistic & $p$ value && Order & $G$ statistic & $p$ value\\ \midrule
%Cluster 1 & 1 & 5.99 & 0.02 && 1 & 0.00 & 0.95 \\
%Cluster 2 & 4& 1.616& 0.20&& 4& 3.4 & 0.02\\
%Cluster 3& 1& 1.61& 0.21&& 1& 0.39 & 0.54\\

%\hline
%& \multicolumn{6}{c}{Covid-19 Florida}\\
%\toprule
%& \multicolumn{3}{c}{Cases} & \phantom{abc}& \multicolumn{3}{c}{Media} &
%\phantom{abc}\\
%\cmidrule{2-4} \cmidrule{6-8} \cmidrule{10-12}
%& Order & $G$ statistic & $p$ value && Order & $G$ statistic & $p$ value\\ \midrule
%Cluster 1 & 1 & 0.40 & 0.53&& 1 &2.93 & 0.1 \\
%Cluster 2 & 1& 0.066& 0.81&& 1& 0.69& 0.41\\
%Cluster 3& 2& 1.13& 0.34&& 1& 0.02& 0.88\\

%\hline
%& \multicolumn{6}{c}{Covid-19 Texas}\\
%\toprule
%& \multicolumn{3}{c}{Cases} & \phantom{abc}& %\multicolumn{3}{c}{Media} &
%\phantom{abc}\\
%\cmidrule{2-4} \cmidrule{6-8} \cmidrule{10-12}
%& Order & $G$ statistic & $p$ value && Order & $G$ statistic & $p$ value\\ \midrule
%Cluster 1 & 1 & 0.09 & 0.77 && 1 &0.12 & 0.89 \\
%Cluster 2 & 1& 0.236& 0.63&& 1& 0.99& 0.45\\
%Cluster 3& 1& 0.24& 0.63&& 4& 0.03& 0.87\\

\bottomrule
\end{tabular}
\caption{%Results of a Granger-causality test on whether confirmed cases or media preceded each cluster. In all cases H0: Cases/media does not Granger-cause cluster.%
Results of a Granger-causality test on whether confirmed cases (left-side columns) or media (right side columns) preceded the centroids of each cluster. We only find evidence of Granger-causality in the case of flu media mentions and cluster 2. In the table, "Order" represents the number of lagged weeks.}
\end{table}

\bibliography{sample.bib}
\printindex